\documentclass{article}
\usepackage[accepted]{icml2019_arX}

\usepackage{graphicx,amsthm} 
\usepackage{subfigure}
\usepackage{wrapfig}
\usepackage{sidecap}
\usepackage{multicol}

\usepackage{bm}
\usepackage{amsmath}
\usepackage{amssymb}
\usepackage{amsfonts}
\usepackage{latexsym}
\usepackage{multirow,tikz,amsmath,comment}
\usepackage{enumitem}
\usepackage[utf8]{inputenc} 
\usepackage[T1]{fontenc}    
\usepackage{hyperref}       
\usepackage{url}            
\usepackage{booktabs}       
\usepackage{amsfonts}       
\usepackage{nicefrac}       
\usepackage{microtype}      
\usepackage{sidecap}
\usepackage{natbib}

\newcommand\independent{\protect\mathpalette{\protect\independenT}{\perp}}
\def\independenT#1#2{\mathrel{\rlap{$#1#2$}\mkern2mu{#1#2}}}

\newtheorem{theorem}{Theorem}
\newtheorem{corollary}{Corollary}
\newtheorem{lemma}{Lemma}

\begin{document}
	
	\twocolumn[
	\icmltitle{Causal Discovery and Forecasting in Nonstationary Environments with State-Space Models}
	
	\icmlsetsymbol{equal}{*}
	
	\begin{icmlauthorlist}
		\icmlauthor{Biwei Huang}{goo}
		\icmlauthor{Kun Zhang}{goo}
		\icmlauthor{Mingming Gong}{goo,ed}
		\icmlauthor{Clark Glymour}{goo}
	\end{icmlauthorlist}
	
	\icmlaffiliation{goo}{Department of Philosophy, Carnegie Mellon University, Pittsburgh}
	\icmlaffiliation{ed}{Department of Biomedical Informatics, University of Pittsburgh, Pittsburgh}
	
	\icmlcorrespondingauthor{Biwei Huang}{biweih@andrew.cmu.edu}
	\icmlcorrespondingauthor{Kun Zhang}{kunz1@cmu.edu}

	\icmlkeywords{Causal Discovery, Nonstationarity, Forecasting, Nonlinear State Space Model}
	
	\vskip 0.3in
	]
	
	\printAffiliationsAndNotice{} 

%
	\begin{abstract}
	  In many scientific fields, such as economics and neuroscience, we are often faced with nonstationary time series, and concerned with both finding causal relations and forecasting the values of variables of interest, both of which are particularly challenging in such nonstationary environments. In this paper, we study causal discovery and forecasting for nonstationary time series. By exploiting a particular type of state-space model to represent the processes, we show that nonstationarity helps to identify causal structure and that forecasting naturally benefits from learned causal knowledge. Specifically, we allow changes in both causal strengths and noise variances in the nonlinear state-space models, which, interestingly, renders both the causal structure and model parameters identifiable. Given the causal model, we treat forecasting as a problem in Bayesian inference in the causal model, which exploits the time-varying property of the data and adapts to new observations in a principled manner.  Experimental results on synthetic and real-world data sets demonstrate the efficacy of the proposed methods.
	\end{abstract}

	\section{Introduction}
	 
	 
	 One of the fundamental problems in empirical sciences is to make prediction for passively observed data (a task that machine learning is often concerned with) or to make prediction under interventions. In order to make prediction under interventions, one has to find and make use of causal relations. Discovering causal relationships from observational data, known as causal discovery, has recently attracted much attention. In many scientific fields, we are often faced with nonstationary time series, and concerned with both finding causal relations and forecasting the values of variables of interest, both of which are particularly challenging in such nonstationary environments. 
	 
	 
	Traditional methods of causal discovery usually focus on independent and identically distributed (i.i.d.) data or stationary processes, and assume that the underlying causal model is fixed. Such methods include constraint-based methods \cite{SGS93}, score-based methods \cite{Chickering02,Heckerman95,Huang18_KDD}, and functional causal model-based approaches \cite{Shimizu06,Zhang06_iconip,Hoyer09,Zhang_UAI09}. Specifically, constraint-based methods and score-based methods recover the causal graph up to the Markov equivalence class, within which some causal directions may not be identifiable. 
	Presuming certain constraints on the class of causal mechanisms, functional causal model-based approaches exploit asymmetries between causal and anti-causal directions.
	 
	 Those traditional methods may not be practical in a number of situations. The assumption of a fixed causal model may not hold in practice, especially for time series, where the underlying data generating processes may change over time. For example, neural connectivity in the brain may change over time or across different states. The influences between macroeconomic variables may be affected by latent common factors, e.g., economic policies, which may change across different time periods and contribute to nonstationarity of observed macroeconomic variables. If we directly apply causal discovery methods which are designed for a fixed causal model, they may give us misleading results, e.g., spurious edges and wrong causal directions; see e.g., \citet{Zhang17_IJCAI}. A second issue is that, with functional causal model-based approaches, there are cases where causal directions are not identifiable, such as the linear-Gaussian case and the case with a general functional class. Hence, this criterion for direction identification is not generally applicable \cite{Zhangetal15_TIST}. Therefore, it is beneficial to investigate other asymmetric criteria for the purpose of causal discovery.
	
     Interestingly, several research papers have shown that nonstationarity contains useful information for causal discovery \cite{Hoover90, Tian01, Huang15, Zhang17_IJCAI,Huang17_ICDM, invariantPred,  Amir18_NIPS}. Nonstationarity may result from a change in the underlying mechanisms, which is related to soft intervention \cite{Intervention1} in the sense that both result in probability distribution changes, while nonstationarity can be seen as a consequence of soft interventions done by nature.
	 Furthermore, from a causal view, it has been postulated that if there is no confounder, the marginal distribution $P$(cause) and the conditional distribution $P$(effect|cause) represent independent mechanisms of nature \cite{Pearl00, Janzing10}, which is related to the exogeneity notion \cite{Engle83_exogeneity, Zhang15_exogeneity}. How to characterize such an independence or exogeneity condition is an issue. Thanks to nonstationarity, the independence between probability distributions can be characterized statistically; in the causal direction, the causal modules $P$(cause) and $P$(effect|cause) change statistically independently, while $P$(effect) and $P$(cause|effect) change dependently generically.
	
	 On the other hand, forecasting from nonstationary data is usually hard. 
	 In this paper, we argue that forecasting can benefit from causal knowledge for
	 nonstationary processes.
     First, from the causal view, the distribution shift in nonstationary data is usually constrained--it might be due to the changes in data generating processes of only a few variables. By detecting these key variables, we only need to update the distributions corresponding to these variables. In complex models, the savings can be enormous; a reduction in the number of modeling variables can translate into substantial reduction in the sample complexity. 
	 Second, by making use of the information from causal structure, each causal module changes independently and thus can be considered separately. 
	 The changes in the causal modules are usually simpler (or more natural) than those in conditional distributions that do not represent causal mechanisms, which also reduces the difficulty of prediction. Third, the causal knowledge makes the forecasts more interpretable. We can gain insight into which factors affect the target variable and how to manipulate the system properly.
	
	 In this paper, we study causal discovery and forecasting for nonstationary time series. We provide a principled investigation of how causal discovery benefits from nonstationarity and how the learned causal knowledge facilitates forecasting. Particularly, we formalize causal discovery and forecasting under the framework of nonlinear state-space models. Our main contributions are as follows:
	\begin{itemize}[noitemsep,topsep=0pt]
		\item In Section \ref{Sec: Model}, we formalize a time-varying causal model to represent the underlying causal process in nonstationary time series. We allow changes in both causal strengths and noise variances, as well as changes of causal structure in the sense that some causal influences may vanish or appear over some periods of time. 
		\item In Section \ref{Sec: Identifiability}, we show the identifiability of the proposed causal model under mild conditions; both the causal structure and model parameters are identifiable. 
		\item In Section \ref{Sec: Identification}, we give a way to estimate the proposed causal model. It can be transformed to the task of standard estimation of nonlinear state-space models.
		\item In Section \ref{Sec: Forecasts}, we show that causal models benefit forecasting. Given the causal model, we treat forecasting as a Bayesian inference problem in the causal model, which exploits the time-varying property of the data and adapts to new observations in a principled manner. 
	\end{itemize}


\section{Motivation and Related Work}

Identification of causal relationships from observational data is attractive for the reason that traditional randomized experiments may be hard or even impossible to do. Over the past decades, prominent progress has been made in this area. Constraint-based methods use statistical tests (conditional independence tests) to find causal skeleton and determine orientations up to the Markov equivalence class; widely-used methods include PC and FCI \cite{SGS93}. Score-based methods define a score function that measures how well an equivalence class fits the observed data and search through possible equivalence classes to find the best scored one \cite{Heckerman95,Chickering02,Huang18_KDD}. 
It was later shown that with functional causal model-based approaches, it is possible to recover the whole causal graph with certain constraints on the functional class of causal mechanisms, by making use of asymmetries between causal and anti-causal directions. For example, in the case of linear causal relationships, the non-Gaussianity of noise terms helps to identify the causal direction; in the causal direction, the noise term is independent of hypothetical causes, while independence does not hold in the anti-causal direction. For instance, the linear non-Gaussian acyclic model (LiNGAM) \cite{Shimizu06} uses this property for causal discovery. 

Granger causality \cite{Granger69} is widely applied in time series analysis, especially in economics. It concerns time-lagged relationships and assumes that the underlying causal strengths and noise variances are fixed. A more recent method based on structural vector-autoregressive models further incorporates contemporaneous causal relationships \cite{SVAR10}. However, these methods are only appropriate for stationary time series, while in real-world problems, it is commonplace to encounter nonstationary data. If we directly apply the above approaches to nonstationary data, it may lead to spurious edges or wrong causal directions; see e.g., \citet{Zhang17_IJCAI}.


More recently, causal discovery methods for nonstationary data have been proposed \cite{Tian01, invariantPred, Zhang17_IJCAI}. In particular, in \citet{Zhang17_IJCAI}, it adds a surrogate variable, e.g., time or domain index, to the causal system to account for changing causal relations and to determine causal directions by exploiting the independent change between $P$(cause) and $P$(effect|cause). Particularly, it uses kernel distribution embeddings to describe shifting probabilistic distributions in a non-parametric way. Despite its general applicability in theory, in practice, it may be limited in several aspects. With kernels, the computational complexity is $O(N^3)$, where $N$ is the sample size, which is expensive and makes it intractable in large data sets. Moreover, in practice, it is not easy to choose an appropriate kernel width, and the kernel width can heavily affect the results.


Another set of studies have tried to model time-varying relationships - such relationships are either not necessarily causal, or causal relationships in which the causal direction is already known in advance, e.g., one can assume that past causes future without contemporaneous causal relationships. Hence, they do not have the phase of discovering causal structure from observational data. For the former case, representative work includes the estimation of time-varying precision matrix by minimizing the temporally smoothed $L_1$ penalized regression \cite{Kolar_12}. For the latter, it includes research studies in dynamic Bayesian networks \cite{DBN92,Song_NIPS09}. 
However, in practice, it is often the case that some causal interactions occur in the same time period, and thus it is important to consider contemporaneous causal relations, especially in time series with low temporal resolutions \cite{SVAR10, Subsampling_ICML15}, in aggregated data \cite{Aggre_Gong17}, or in equilibrium data. 

%
%

For forecasting with nonstationary data, basically two types of methods are usually used: active approaches and passive approaches \cite{Nonsta_pred1, Nonsta_pred2}. Specifically, the active approach updates the model only when a change is detected, which limits its applicability for time series with gradual changes. The passive approach, such as the dynamic linear model, does not actively detect the drift in environments, but performs a continuous adaptation of the model every time new data arrive. 

To the best of our knowledge, the present paper is the first work on simultaneous causal discovery (covering both contemporaneous and time-lagged causal relationships) and forecasting in nonstationary environments, where forecasting directly benefits from causal modeling in a natural way. 

\section{Time-Varying Linear Causal Models} \label{Sec: Model}
Suppose that we have $m$ observed time series $X_t = \big ( x_{1,t}, \cdots, x_{m,t} \big )^{\text{T}}$, satisfying the following generating process:
\begin{equation} \label{FCM}
x_{i,t} = \sum_{x_j \in \mathbb{PA}_i} b_{ij,t} x_{j,t} + e_{i,t}, 
\end{equation}
where $\mathbb{PA}_i$ is the set of direct instantaneous causes of $x_{i}$, $x_j \in \mathbb{PA}_i$ is the $j$th direct cause of $x_{i}$, $b_{ij,t}$ is the causal coefficient from $x_{j,t}$ to $x_{i,t}$, and $e_{i,t}$ is the Gaussian noise term with $e_{i,t} \sim \mathcal{N}(0, \sigma_{i,t}^2)$, which indicates influences from unmeasured factors. The noise distribution does not have to be Gaussian; here we make this assumption mainly for the purpose of showing that even when causal relationships are linear, and the noise terms are Gaussian, the causal model is still identifiable by using nonstationarity. In real-world problems, other appropriate noise distributions can be applied. We will see in Section \ref{Sec: Identifiability} that the identifiability of the time-varying causal model does not require the Gaussian assumption.


The causal process is assumed to have the following properties.
\begin{itemize}[noitemsep,topsep=0pt]
	\item Let $B_t$ be the $m \times m$ causal adjacency matrix with entries $b_{ij,t}$, and denote by $G_t$ the corresponding binary matrix (quantitative causal adjacency matrix), with $G_t(j,i)=1$ if and only if $b_{ij,t} \neq 0$ and zero otherwise. We assume that the graph union $G = G_1 \cup \cdots \cup G_T$ is acyclic.
	\item We allow each causal coefficient $b_{ij,t}$ and noise variance $\sigma_{i,t}^2$ to change over time and model the changes by the following autoregressive models:
	\begin{equation} \label{latent_equation}
	\begin{array}{lll}
	b_{ij,t} = \alpha_{ij,0} + \sum\limits_{p=1}^{p_l} \alpha_{ij, p} b_{ij,t-p} + \epsilon_{ij,t}, \\
	h_{i,t} = \beta_{i,0}  + \sum\limits_{q=1}^{q_l}  \beta_{i,q} h_{i,t-q} + \eta_{i,t}, 
	\end{array}
	\end{equation}
	respectively, where $ \epsilon_{ij,t} \sim \mathcal{N}(0,w_{ij})$, $\eta_{i,t} \sim \mathcal{N}(0,v_i)$, and $h_{i,t} = \log(\sigma^2_{i,t}) $ models the volatility of the observed time series. Each causal coefficient and log-transformed noise variance changes independently. Again here the distributions of $\epsilon_{ij,t}$ and $\eta_{i,t}$ are not necessarily Gaussian; for example, we can easily extend it to mixture of Gaussian distributions. Note that this formulation includes the case where only causal coefficients change with time, while noise distributions stay constant, i.e., $e_{i,t} \sim \mathcal{N}(0, \sigma_{i}^2)$, $\forall i \in \mathbf{N}^+$.
	\item We allow changes in causal structure that some causal edges may vanish or appear over some periods of time.
\end{itemize}

\begin{figure}[htp!]
 	\setlength{\abovecaptionskip}{10pt}
	\setlength{\belowcaptionskip}{10pt}
 	\vspace{-.3cm}
	\centering
	\includegraphics[width = .4\textwidth ]{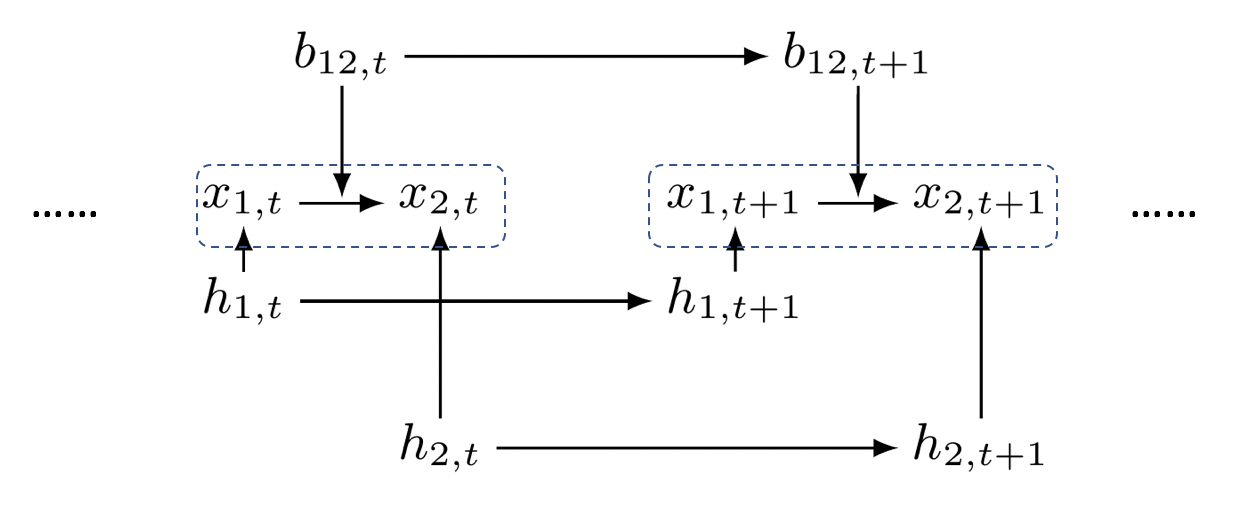}
	\caption{A graphical representation of generating processes of the time-varying causal network.}
	\label{fig:generate_process}
\end{figure}

Equation (\ref{FCM}) can be represented in the matrix form, with 
\begin{equation} 
\label{FCM_matrix}
X_t = (I_m - B_t)^{-1} E_t,
\end{equation}
where $I_m$ is an $m \times m$ identity matrix, and $E_t = \big ( e_{1,t}, \cdots, e_{m,t} \big )^{\text{T}}$.
Thus, by combining causal process (\ref{FCM_matrix}) and autoregressive functions (\ref{latent_equation}), we have the following causal model:
\begin{equation} 
\label{SSM}
\left\{  
\begin{array}{lll}
X_t & = (I_m - B_t)^{-1} E_t, \\
b_{ij,t} & = \alpha_{ij,0} + \sum\limits_{p=1}^{p_l} \alpha_{ij,p} b_{ij,t-p} + \epsilon_{ij,t}, \\  
h_{i,t} & = \beta_{i,0} + \sum\limits_{q=1}^{q_l} \beta_{i q} h_{i,t-q} + \eta_{i,t},  \\
\end{array}
\right.
\end{equation}
with $\epsilon_{ij,t} \sim \mathcal{N}(0,w_{ij})$ and $\eta_{i,t} \sim \mathcal{N}(0,v_i)$, for $t = \max(p_l,q_l),\cdots,T$.
Figure \ref{fig:generate_process} gives the graphical representation of generating processes of the time-varying causal network.

In real-world problems, there may also exist time-delayed causal relations. To consider both time-delayed and instantaneous causal relations, we modify equation (\ref{FCM}) to
\begin{equation} 
x_{i,t} = \sum_{x_j \in \mathbb{PA}_i} b_{ij,t} x_{j,t} + \sum_{s=1}^{s_l} \sum_{x_k \in \mathbb{PL}_i} c_{ik,t}^{(s)} x_{k,t-s} + e_{i,t},
\end{equation} 
where $\mathbb{PL}_i$ is the set of lagged causes of $x_i$, and $c_{ik}^{(s)}$ represents the $s$-lagged causal strength from $x_k$ to $x_i$. Similarly, we model the time-varying lagged causal strength with an autoregressive model,
\begin{equation}
 \begin{array}{ll}
    c_{ij,t}^{(s)} = \gamma_{ij,0}^{(s)} + \sum\limits_{r=1}^{r_l} \gamma_{ij,r}^{(s)} c_{ij,t-r}^{(s)} + \nu_{ij,t}^{(s)},
 \end{array}
\end{equation}
with $\nu_{ij,t}^{(s)} \sim \mathcal{N}(0,u_{ij}^{(s)})$.

In the next section, we are mainly concerned with the identifiability of instantaneous causal relations, while the results are also extended to handle the above delayed causal relations. 

It is worth noting that although the model (\ref{FCM}) is linear in the processes, it is actually nonlinear in the latent processes $b_{ij}$ and $h_i$. Therefore, the time-varying linear causal model is actually a specific type of nonlinear state-space model with respect to hidden variables $b_{ij}$ and $h_i$.  In fact, in Section \ref{Sec: Identification}, we will estimate the proposed model by extending methods for estimating nonlinear state-space models.


%
%

\section{Model Identifiability} \label{Sec: Identifiability}
We show in Theorem \ref{Theorem: identifiability} that the proposed causal model, including causal structure and model parameters, is identifiable under the following conditions:
\begin{itemize}[noitemsep,topsep=0pt]
	\item The underlying instantaneous causal structure is acyclic.
   \item Each causal coefficient varies with time and follows an autoregressive model, and distributions of $e_{i,t}$ are fixed.
\end{itemize}
Note that for identifiability, we do not require the additive noise terms to be Gaussian. Furthermore, we do not assume faithfulness \cite{SGS93}, which is commonly assumed in traditional constraint-based causal discovery. 

\begin{theorem}
	
	Suppose the observed time series, $X_t = \big ( x_{1,t}, \cdots, x_{m,t} \big )^{\text{T}}$, were generated by
	\begin{equation} \label{SSM_T1}
	\left \{
	\begin{array}{llll} 
	x_{i,t}  =  \sum_{x_j \in \mathbb{PA}_i} b_{ij,t} x_{j,t} + e_{i,t},  \\ 
	b_{ij,t}  =  \alpha_{ij,0} + \alpha_{ij,1} b_{ij,t-1} + \epsilon_{ij,t}, 
	\end{array}\right.
	\end{equation}
	where $x_{j,t}$ is the cause of $x_{i,t}$, and $b_{ij,t}$ is the corresponding causal coefficient from $x_{j,t}$ to $x_{i,t}$, which satisfies a first-order autoregressive model with $\alpha_{ij,0}, \alpha_{ij,1} \in (-1,1)$. The additive error, $e_{i,t}$, represents a stationary zero-mean white noise process, i.e.,  $E[e_{i,t}] = 0$, $E[e_{i,t} e_{i,t'}] = \sigma_i^2 \delta_{tt'}$, and $E[e_{i',t} e_{i,t}] = \sigma_i^2 \delta_{ii'}$, where $\sigma_i^2 < \infty$ and $\delta_{tt'}$ is the delta function. 
	Similarly, the error in the autoregressive model of $b_{ij,t}$ satisfies $E[\epsilon_{ij,t}] = 0$ and $E[\epsilon_{ij,t} \epsilon_{ij,t}] = w_{ij}$. In addition, the underlying instantaneous causal structure over $X_t$ is assumed to be acyclic.
	
	Then the model in (\ref{SSM_T1}) is identifiable, including the causal order between $x_{i}$'s and model parameters, when time series are long enough.
	\label{Theorem: identifiability}
\end{theorem}

Here we give a sketch of the proof. For complete proofs of the theoretical results reported in the paper, please refer to the supplementary material.	
\begin{proof}[Proof sketch]
	\begin{enumerate}[topsep=0pt] 
		\item First identify the root cause.\\		
		Let
		\begin{equation*}
		S(t,t+p)_i := E[x_{i,t}^2  x_{i,t+p}^2].
		\end{equation*}
		Let $r_0$ be the index of the root cause, and $\mathbf{V}_s = \mathbf{V} \backslash r_0$ denote the indices of the remaining processes, with $\mathbf{V}  = \{1,\cdots,m\}$. Then we will have
		\begin{equation*}
		\begin{array}{ll}
		S(t,t+p)_{r_0} - S(t,t+p-1)_{r_0} = 0;\\
		S(t,t+p)_{r_s} - S(t,t+p-1)_{r_s} < 0, \quad \forall r_s \in \mathbf{V}_s.
		\end{array}
		\end{equation*}
		 The reason is that the root cause does not receive changing influences from other processes.		
		For the root cause, $S(t,t+p)_{r_0} = \sigma_{r_0}^4$, where $\sigma_{r_0}^2$ is the noise variance in the causal model of $x_{r_0}$, so we can also identify the noise variance of $x_{r_0}$.
		
		\item Next, iteratively identify the remaining causal graph.\\		
		Suppose that we have identified $n$ processes that are the earliest according to the causal order. We then identify the next variable according to the causal order. Let $\mathbf{V}_n$ represent variable indices of the first $n$ processes and let $\mathbf{V}_{\tilde{n}} = \mathbf{V} \backslash \mathbf{V}_n$. For any $r_s \in \mathbf{V}_{\tilde{n}}$, we show that if and only if $x_{r_s}$ is the next according to the order, $S(t,t+p)_{r_s}$ is a linear combination of cross-statistics of different orders of $x_{\mathbf{V}_n,t}$. In this way, we can identify the causal graph of the first $n+1$ processes. In addition, the corresponding parameters are also identifiable, according to the identifiability of the varying coefficient regression models \cite{Identifiability_SSM}.

		Repeating this procedure until we go through all processes, we have the identifiability of the whole causal model.
		
	\end{enumerate}	
\end{proof}

It is easy to extend the above identifiability result to the case when there are both time-lagged and instantaneous causal relations, which is given in Corollary \ref{Corollary: identifiability}, since for lagged causal relations, their causal directions are fixed (from past to future), and thus, it reduces to a parameter identification problem.
\begin{corollary}
	
	Suppose that the $m$ observed time series, $X_t = \big ( x_{1,t}, \cdots, x_{m,t} \big )^{\text{T}}$, satisfy the following generating process:
   \begin{equation} \label{SSM_T2}
	\! \! \! \! \! \! \left \{
	\begin{array}{lll} 
\! \! \! \!	x_{i,t} = \! \! \! \! \sum \limits_{x_j \in \mathbb{PA}_i} b_{ij,t} x_{j,t} + \sum  \limits_{s=1}^{s_l} \sum  \limits_{x_k \in \mathbb{PL}_i} c_{ik,t}^{(s)} x_{k,t-s} + e_{i,t},\\
\! \! \! 	b_{ij,t}  = \alpha_{ij,0} + \alpha_{ij,1} b_{ij,t-1} + \epsilon_{ij,t}, \\
\! \! \! c_{ij,t}^{(s)}  = \gamma_{ij,0}^{(s)} + \sum\limits_{r=1}^{r_l} \gamma_{ij,r}^{(s)} c_{ij,t-r}^{(s)} + \nu_{ij,t}^{(s)},
	\end{array}\right.
	\end{equation}
	where $c_{ij,t}^{(s)}$ represents the $s$-lagged causal coefficient, which satisfies an autoregressive model with $\gamma_{ij,0}^{(s)}, \gamma_{ij,r}^{(s)}  \in (-1,1)$. The additive error, $\nu_{ij,t}^{(s)}$, represents a stationary zero-mean white noise process,with $E[\nu_{ij,t}^{(s)}] = 0$ and $E[\nu_{ij,t}^{(s)} \nu_{ij,t}^{(s)}] = u_{ij}^{(s)}$. Other notations $b_{ij,t}$, $e_{i,t}$, $\alpha_{ij,0}$, $\alpha_{ij,1}$, and $\epsilon_{ij,t}$ are the same as in Theorem \ref{Theorem: identifiability}. In addition, the underlying instantaneous causal structure over $X_t$ is assumed to be acyclic.

	Then the model in (\ref{SSM_T2}) is identifiable, including the causal order between $x_{i}$'s and model parameters, when time series are long enough.
\label{Corollary: identifiability}
\end{corollary}



The above results do not take into account the changeability of $\sigma_{i}^2$. For the general case where $\sigma_{i}^2$ futher changes, our empirical results strongly suggest that the causal model is also identifiable, although currently there is no straightforward, concise proof for it.

 
\section{Model Identification}  \label{Sec: Identification}
The model defined in equation (\ref{SSM}) can be regarded as a nonlinear state-space model, with causal coefficients and log-transformed noise variances being latent variables $Z = \big \{ \{b_{ij}\}_{i,j=1}^m, \{h_i\}_{i=1}^m \big \}$, and model parameters $\theta = \big \{ \{\alpha_{ij,p}\}, \{\beta_{i,q}\}, \{w_{ij}\}, \{v_i\} \big \}$. Therefore, it can be transformed to a nonlinear state-space model estimation problem. In particular, we exploit an efficient stochastic approximation expectation maximization (SAEM) algorithm \cite{SAEM}, combined with conditional particle filters with ancestor sampling (CPF-AS) in the E step \cite{CPF, SAEM_CPF}, for model estimation.

\subsection{SAEM Algorithm}
For a traditional EM algorithm, the procedure is initialized at some $\theta_0 \in \Theta$ and then iterates between two steps, expectation (E) and maximization (M):
	\vspace{-0.1cm}
\begin{itemize}[noitemsep,topsep=0pt]
	\item [(E)] Compute $p_{\theta^{k-1}} (Z|X)$ and the lower bound of the log-likelihood, $\mathcal{Q}(\theta, \theta^{k-1})$, with 
	\vspace{-0.1cm}
	\begin{equation*}
	\mathcal{Q}(\theta, \theta^{k-1}) = \int p_{\theta^{k-1}} (Z|X) \log p_{\theta} (Z, X) \, d Z.
	\end{equation*}
	\item [(M)] Compute $\theta^{k} = \arg \max_{\theta \in \Theta} \mathcal{Q}(\theta, \theta^{k-1})$.
\end{itemize}
In the E step, we need to compute the expectation under the posterior $p_{\theta^{k-1}} (Z|X)$, which is intractable in our case, since $p(X,Z)$ is not Gaussian. To address this issue, SAEM computes the E step by Monte Carlo integration and uses a stochastic approximation update of the quantity $\mathcal{Q}$:
\begin{equation}
\resizebox{1.02\hsize}{!}{$ \! \! \tilde{\mathcal{Q}}_k (\theta)  \! = \! (1 - \lambda_k)  \tilde{\mathcal{Q}}_{k-1} (\theta) \! + \! \lambda_k  \! \sum \limits_{j=1}^{M} \frac{\omega_T^{(k,j)}}{\sum_l \omega_T^{(k,l)}} \log p_{ \theta} (X_{1 : T },  \mathring{Z}_{1:T}^{(k,j)}), $}
 \label{E'}
\end{equation}
	\vspace{-0.1cm}
where $\mathring{Z}$ indicates sampled particles of $Z$, $\omega_T^{(k,j)}$ the weight of $j$th particle at $k$th iteration, $M$ the generated number of particels, $X_{1:T} = \{X_t\}_{t=1}^T$, $\mathring{Z}_{1:T}^{(k,j)} = \{\mathring{Z}_t^{(k,j)}\}_{t=1}^T$, and $\{\lambda_k\}_{k \geq 1}$ is a decreasing sequence of positive step size, with $\sum_{k} \lambda_k = \infty$ and $\sum_{k} \lambda_k^2 < \infty$.
The E-step is thus replaced by the following:
\begin{itemize}[noitemsep,topsep=0pt]
	\item [(E$'$)] At each iteration, generate $M$ particles of $\mathring{Z}_{1:T}^{(k,j)}$ from $p_{\theta^{k-1}} (Z|X)$ and compute $ \tilde{\mathcal{Q}}_k (\theta)$ according to (\ref{E'}). (A method for sampling from $p_{\theta^{k-1}} (Z|X)$ is introduced in the next section.)
\end{itemize}

Under appropriate assumptions, SAEM is shown to converge for fixed $M$, as $k \rightarrow \infty$ \cite{SAEM}. The model parameters in the M step are updated by setting $\frac{\partial \tilde{\mathcal{Q}}_k (\theta) }{\partial \theta} = 0$. The detailed derivations are given in Section $S3$ in supplementary materials.
The computational complexity in each iteration is $O(m^3 \times M \times T)$, where $m$ is the number of variables, $M$ the number of sampled particles (we used $M=15$), and $T$ the length of time series. 

\subsection{Conditional Particle Filter with Ancestor Sampling}
To sample particles $\mathring{Z}$ from the posterior distribution, we use conditional particle filtering with ancestor sampling (CPF-AS) \cite{SAEM_CPF}. The CPF-AS procedure is a sequential Monte Carlo sampler, akin to a standard particle filter but with the difference that one particle at each time step is specified as a priori. Let these prespecified particles be $\mathring{Z}'_{1:T} = \{\mathring{Z}'_t\}_{t=1}^T$. Let $\{\mathring{Z}_{1:t-1}^{(j)}, \omega_{t-1}^{(j)}\}_{j=1}^M$ be a weighted particle system targeting $p_{\theta}(\mathring{Z}_{1:t-1} | X_{1:t-1})$. To propagate this sample to time $t$, we introduce the auxiliary variable $s_t^j$, referred to the ancestor particle of $\mathring{Z}_t^{(j)}$. To generate a specific particle $\mathring{Z}_t^{(j)}$ at time t, we first sample the ancestor index with $P(s_t^j = i) \propto \omega_{t-1}^i$. Then $\mathring{Z}_t^{(j)}$ is sampled from $\mathring{Z}_t^{(j)} \sim f_{\theta}(\mathring{Z}_t | \mathring{Z}_{t-1}^{s_t^j}), \text{ } j = 1,\cdots,M\!-\!1$.
The $M$th particle is sampled deterministically: $\mathring{Z}_t^{(M)} = \mathring{Z}'_t$. We sample the ancestor index $s_t^M$ with $P(s_t^M = j) \propto \omega_{t-1}^{(j)} f_{\theta}(\mathring{Z}'_t | \mathring{Z}_{t-1}^j)$.
Finally, all the particles are assigned importance weights, $\omega_t^{(j)} = W_{\theta,t}(\mathring{Z}_t^{(j)}, \mathring{Z}_{t-1}^{s_t^j})$, where the weight function is given by $p_{\theta}(X_t | \mathring{Z}_t)$. The CPF-AS is summarized in Algorithm $S1$ (Section $S4$) in supplementary materials.

 \subsection{Causal Graph Determination}
 The causal graph is determined from the sampled particles. 
 With finite samples, there exist estimation errors; for example, even when there is no causal edge from $x_j$ to $x_i$, the estimation $\hat{b}_{ij,t}$ may not be exactly zero but some small values. To determine whether there is a causal edge from $x_j$ to $x_i$, we check both the mean and the variance of $\hat{b}_{ij,t}$. 
Specifically, if both $\bar{\hat{b}}_{ij} =  \frac{1}{T} \sum_{t=1}^{T} \hat{b}_{ij,t} < \alpha$ and $\frac{1}{T} \sum_{t=1}^{T} (\hat{b}_{ij,t}-\bar{\hat{b}}_{ij})^2 < \alpha$, we determine that there is no causal edge from $x_j$ to $x_i$, where $\alpha$ is a threshold. 

  In our model, we estimate the causal adjacency matrix $B_t$ directly. 
  Recall that LiNGAM \cite{Shimizu06} first estimates $A = (I-B)^{-1}$ and then recover the underlying adjacency matrix $B$ by performing extra permutation and rescaling, since $W$ is only identified up to permutation and scale. We directly model the causal process, represented by $B_t$, with the following advantages:
  \begin{itemize}[noitemsep,topsep=0pt]
  	\item It is easy to add prior knowledge of causal connections. In practice, experts may have domain knowledge about some causal edges.
  	\item One can directly enforce sparsity constraints on the causal adjacencies; even if $B_t$ is sparse, $(I-B_t)^{-1}$ is not necessarily sparse, so enforcing the sparsity of causal adjacency would be more difficult when working with $A$. Section $S5$ in the supplementary materials explains how to add sparsity constraints on causal adjacency matrix $B_t$ and on $b_{ij,t} - b_{ij,t-1}$, which ensures smooth changes of $b_{ij,t}$ over time.
  	\item The estimation procedure directly outputs the causal adjacency matrix, without additional steps of permutation and rescaling, which are usually expensive.
  \end{itemize}
  
\vspace{-0.1cm}
\section{Forecasting with Time-Varying Causal Models} \label{Sec: Forecasts}
\vspace{-0.1cm}
   After identifying the causal model, we aim to do forecasting by taking advantage of the causal information.
   Suppose that we have observational data  $\tilde{X}_{1:T+1}$ and $Y_{1:T}$, with $X = \{\tilde{X},Y\}$, and we want to predict $Y_{T+1}$. We denote the Markov blanket of $Y$ by $\mathcal{M}_Y = \mathcal{P}_Y \cup \mathcal{C}_Y \cup \mathcal{S}_Y$, where $\mathcal{P}_Y$ denotes the set of parents of $Y$, $\mathcal{C}_Y$ the set of children of $Y$, and $\mathcal{S}_Y$ the set of spouses of $Y$. Given its Markov blanket, $Y$ is independent of remaining variables in the causal system; thus, $\mathcal{M}_Y$ contains all the information that is needed to predict $Y$. The posterior of $Y_{T+1}$ given its Markov blanket at time $T+1$ can be represented as
	\begin{equation}
	\begin{array}{ll}
	  \! \! \! \! \! \! \! & p(Y_{T+1} | \mathcal{M}_{Y,T+1}) \\
	    \! \!  \! \! \! \! \! \propto  \! \! \! & p(Y_{T+1} | \mathcal{P}_{Y,T+1}) \prod \limits_{\tilde{X}_{C_i} \in \mathcal{C}_Y } p(\tilde{X}_{C_i, T+1} | \mathcal{P}_{C_i, T+1}),
	\end{array}
   \end{equation}
	where $\tilde{X}_{C_i} \in \mathcal{C}_Y $ is the $i$th child of $Y$, and $\mathcal{P}_{C_i} \in \mathcal{M}_Y$ denotes the parents of $\tilde{X}_{C_i}$ in $\mathcal{M}_Y$. Let $\vec{b}_Y$ and $\sigma^2_Y$ denote the corresponding causal coefficients and noise variance in the functional causal model of $Y$. Let $D_T := \{\tilde{X}_{1:T}, Y_{1:T} \}$. Then we have
	\begin{equation}
	\begin{array}{lll}
	& \! p(Y_{T+1} | \mathcal{P}_{Y,T+1})\\
\!\!\!	=  & \! \! \int \int \int \int p(Y_{T+1} | \mathcal{P}_{Y,T+1},\vec{b}_{Y,T+1}, \sigma^2_{Y,T+1})\\ 
& \! \! p(\vec{b}_{Y,T+1} | \vec{b}_{Y,T}) 
	  p(\vec{b}_{Y,T} | D_T)  p( \sigma^2_{Y,T+1} |  \sigma^2_{Y,T})\\
  & \! \! p( \sigma^2_{Y, T} |  D_T) 
	 \,d \vec{b}_{Y,T+1} \,d \vec{b}_{Y,T} \,d \sigma^2_{Y,T+1}  \,d \sigma^2_{Y,T}.
	\end{array}
	\label{Equ:infer1}
	\end{equation}

	Since each coefficient changes independently, $p(\vec{b}_{Y,T+1} | \vec{b}_{Y,T}) $ can be written as
	\begin{equation}
	p(\vec{b}_{Y,T+1} | \vec{b}_{Y,T})= \prod_{b_{Y}^j \in \vec{b}_{Y}} p(b_{Y,T+1}^j | b_{Y,T}^j),    
		\vspace{-.2cm}
	\end{equation}
	where $b_{Y}^j$ is the $j$th entry in $\vec{b}_{Y}$.
	
	Similarly, let $\vec{b}_{C_i}$ and $\sigma_{C_i}^2$ denote the corresponding causal coefficients and noise variance in the causal model of $\tilde{X}_{C_i}$, respectively. Then we have
	\begin{equation}
	\begin{array}{ll}
	  \! \! \!  \! \! \!  & \! \! \! p(\tilde{X}_{C_i, T+1} | \mathcal{P}_{C_i, T+1}) \\
	  \! \! \!  \! \! \!  = &  \! \! \!   \int \int \int \int p(\tilde{X}_{C_i, T+1} | \mathcal{P}_{C_i, T+1},\vec{b}_{C_i,T+1},\sigma_{C_i,T+1}^2) \\
	  \! \! \!  \! \! \!  &  \! \! \!  p(\vec{b}_{C_i,T+1} | \vec{b}_{C_i,T} ) p(\vec{b}_{C_i,T} | D_T ) p(\sigma_{C_i, T+1}^2 | \sigma_{C_i, T}^2) \\ 
	  \! \! \!  \! \! \!  &  \! \! \!  p(\sigma_{C_i, T}^2 | D_T) 
	 \,d \vec{b}_{C_i,T+1} \,d \vec{b}_{C_i,T} \,d \sigma^2_{C_i, T+1}  \,d \sigma^2_{C_i, T},
	\end{array}
	\label{Equ:infer2}
		\vspace{-.2cm}
	\end{equation}
	with
	\begin{equation}
		\vspace{-.2cm}
	p(\vec{b}_{C_i,T+1} | \vec{b}_{C_i,T})= \prod_{b_{C_i}^j \in \vec{b}_{C_i}} p(b_{C_i,T+1}^j | b_{C_i,T}^j),    
		\vspace{-.1cm}
	\end{equation}
	where $b_{C_i}^j$ is the $j$th entry in $\vec{b}_{C_i}$.
	
	Since $p(Y_{T+1} | \mathcal{P}_{Y,T+1})$ and $p(\tilde{X}_{C_i, T+1} | \mathcal{P}_{C_i, T+1})$ are not necessarily Gaussian, the integrations in (\ref{Equ:infer1}) and (\ref{Equ:infer2}) are not given in closed forms. We use Markov chain Monte Carlo (MCMC) to do Bayesian inference; in particular, we use the Metropolis-Hastings algorithm \cite{MonteCarlo_04}.
	
	$ p(Y_{T+1} | \mathcal{P}_{Y,T+1})$ in equation (\ref{Equ:infer1}) is estimated by Monte Carlo integration,
	\begin{equation*}
	p(Y_{T+1} | \mathcal{P}_{Y,T+1}) =  \sum_{j} p(Y_{T+1} | \mathcal{P}_{Y,T+1},\vec{b}_{Y,T+1}^{(j)}, \sigma_{Y,T+1}^{2 (j)}),
	\end{equation*}
	where $\vec{b}_{Y,T+1}^{(j)}$ and $\sigma_{Y,T+1}^{2 (j)}$, $\forall j$, are samples from $p(\vec{b}_{Y,T+1} | \vec{b}_{Y,T})$ and $p( \sigma^2_{Y,T+1} |  \sigma^2_{Y,T})$, respectively. For $\vec{b}_{Y,T}$ and $\sigma^2_{Y,T}$, we use the generated particles by CPF-AS in model estimation. Similarly, $p(\tilde{X}_{C_i, T+1} | \mathcal{P}_{C_i, T+1})$ in equation (\ref{Equ:infer2}) is also estimated by Monte Carlo integration.
	
	The detailed procedure of estimating $p(Y_{T+1} | \mathcal{M}_{Y,T+1})$ by Metropolis-Hastings is given in Algorithm \ref{Algo: MH}, where $q(\cdot)$ is taken as a normal distribution. The first hundred samples are ignored, due to the ``burn-in" period.

\vspace{-0.1cm}
	\begin{algorithm}   	
	\caption{Forecasting of $Y_{T+1}$ by Metropolis-Hastings}
	\begin{algorithmic}[1]
		\STATE  Initialize $Y^{(0)}$.
		\FOR {$i = 1 \text{ to } N$}
		\STATE Propose: $Y^{\text{candi}} \sim q(Y^{(i)} | Y^{(i-1)})$.
		\STATE Acceptance probability:
		\begin{small}
		\begin{flalign*} 
		& \alpha(Y^{\text{candi}} | Y^{(i-1)}) \\
		=& \min \bigg \{1, 
		\frac{q(Y^{(i-1)} | Y^{\text{candi}}) p(Y^{\text{candi}} | \mathcal{P}_{Y,T+1} )  }{q(Y^{\text{candi}} | Y^{(i-1)}) p(Y^{(i-1)} | \mathcal{P}_{Y,T+1} ) } \\
		&  \cdot \frac{ \prod\limits_{\tilde{X}_{C_i} \in \mathcal{C}_Y } p(\tilde{X}_{C_i, T+1} | Y^{\text{candi}}, \mathcal{P}_{C_i, T+1}/Y^{\text{candi}}) }{ \prod\limits_{\tilde{X}_{C_i} \in \mathcal{C}_Y } p(\tilde{X}_{C_i, T+1} | Y^{(i-1)}, \mathcal{P}_{C_i, T+1}/Y^{(i-1)}) } \bigg \}.
		\end{flalign*} 
	\end{small}
		\STATE Take $u \sim \text{Uniform} (0,1)$.	
		\IF {$u < \alpha$}
		\STATE accept the propose: $Y^{(i)} = Y^{\text{candi}}$,
		\ELSE
		\STATE reject the propose: $Y^{(i)} = Y^{(i-1)}$.
		\ENDIF
		\ENDFOR
		\STATE  Output:  $\hat{Y}_{T+1} = \frac{1}{N-100+1} \sum_{i = 100}^{N} Y^{(i)}$.			
	\end{algorithmic}	 
	\label{Algo: MH}
\end{algorithm}

\vspace{-0.3cm}
\section{Experimental Results} \label{Sec: Experiments}

To show the efficacy of the proposed approach for simultaneous causal discovery and forecasting, we apply it to both synthetic and real-world data.

\vspace{-.2cm}
\paragraph{Synthetic Data}
We considered two types of data generating processes:
\begin{itemize}[noitemsep,topsep=0pt]
	\item [(1)] Only causal strengths $b_{ij}$ change over time according to autoregressive models, but noise variances $\sigma^2_i$ are constant over time.
	\item [(2)] Both causal strengths $b_{ij}$ and noise variances $\sigma^2_i$ change over time according to autoregressive models.
\end{itemize}

We randomly generated acyclic causal structures according to the Erdos-Renyi model \cite{Generate_graph} with parameter 0.3. Each generated graph has 5 variables. The parameters were set in the following way: the fixed noise variance $\sigma^2_i \sim \mathcal{U}(0.1,0.5)$, the noise variance of $b_{ij}$'s autoregressive model $w_{ij} \sim \mathcal{U}(0.01,0.1)$, the noise variance of $h_{i}$'s autoregressive model $v_{i} \sim \mathcal{U}(0.01,0.1)$, the coefficient in $b_{ij}$'s autoregressive model $\alpha_{i, p} \sim  \mathcal{U}(0.8,0.998)$, and the coefficient in $h_{i}$'s autoregressive model $\beta_{i, p} \sim  \mathcal{U}(0.8,0.998)$, where $\mathcal{U}(l,u)$ denotes a uniform distribution between $l$ and $u$.  We also considered different sample sizes $T = 500, 1000, 1500, \text{and } 2000$. For each setting (a particular data generating process and a particular sample size), we generated 50 realizations. 

For causal discovery, we identified the causal structure by the proposed method. We compared it with other well-known approaches in causal discovery, including LiNGAM, Causal Discovery from NOnstationary/heterogeneous Data (CD-NOD) \cite{Zhang17_IJCAI}, the minimal change method (MB) \cite{Amir18_NIPS}, and the identical boundaries method (IB) \cite{Amir18_NIPS}. 
CD-NOD estimates the causal skeleton by constraint-based methods over an augmented set of variables and orients the causal direction by using the modularity property: $P$(cause) $\independent$ $P$(effect|cause). Both IB and MC are designed for multi-domain causal discovery in linear systems. 

In our methods, we randomly initialized the parameters and determined the causal graph by using a threshold (we simply used 0.05) for both the mean and variance of $\hat{b}_{ij,t}$; that is, if $\bar{\hat{b}}_{ij} = \frac{1}{T} \sum_{t=1}^{T} \hat{b}_{ij,t} < 0.05$ and $\frac{1}{T} \sum_{t=1}^{T} (\hat{b}_{ij,t} -  \bar{\hat{b}}_{ij})^2 < 0.05$,  we concluded that there is no edge from $x_j$ to $x_i$. For CD-NOD, 
the kernel width was set empirically \cite{Zhang17_IJCAI}, and the significance level was 0.05. Since both IB and MC methods need data from multiple domains, we segmented the data into non-overlapping domains with sample size 100 in each domain.

\begin{figure}[htp!] 
	\setlength{\abovecaptionskip}{20pt}
	\setlength{\belowcaptionskip}{10pt}
	\vspace{-0cm}
	\includegraphics[width = 0.5\textwidth, height = 3.6cm,  trim={1.5cm 1.8cm 1cm 1cm},clip]{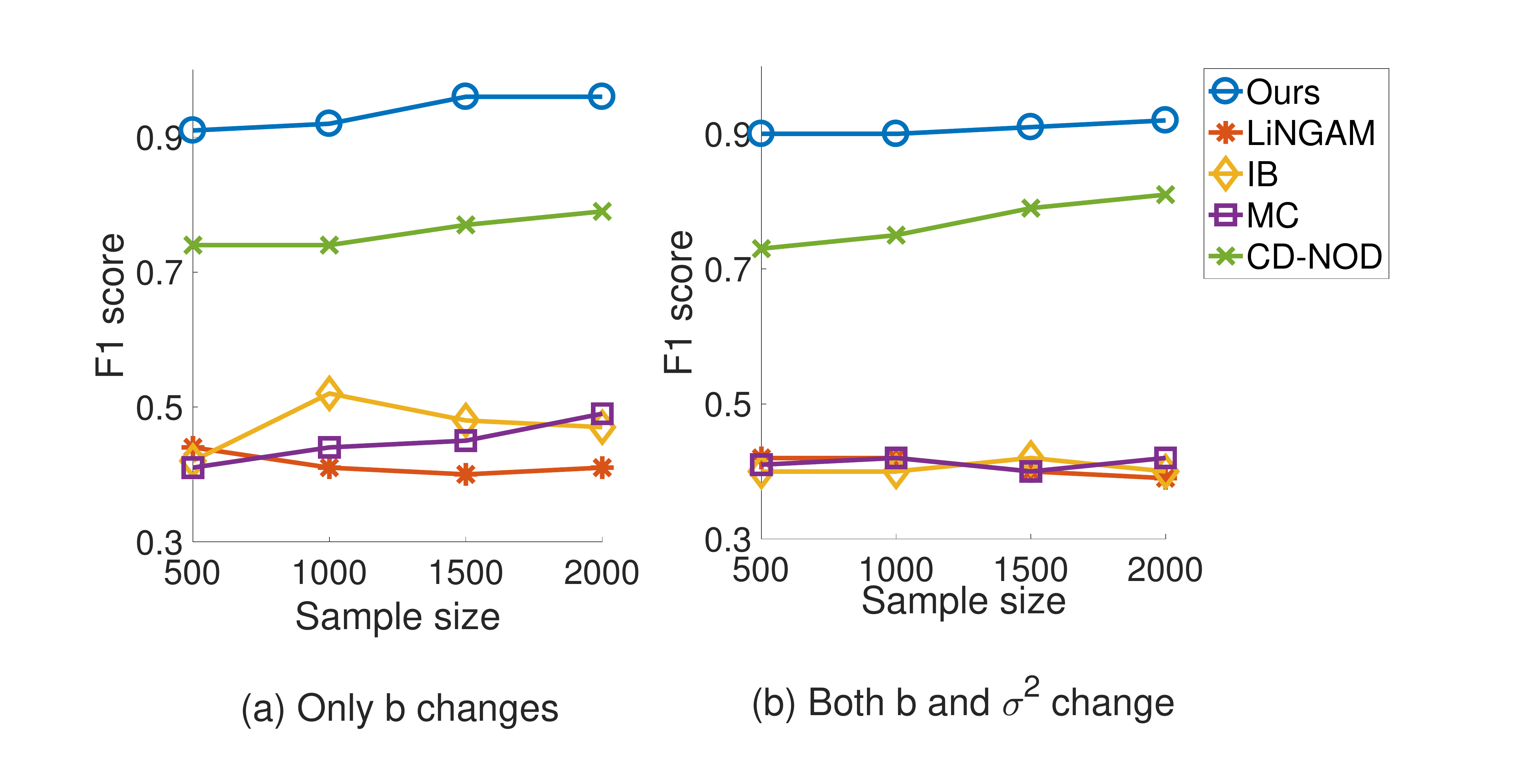} \vspace{-0.7cm}
	\caption{F1 score of the estimated causal graph when (a) only $b_{ij,t}$ changes and when (b) both $b_{ij,t}$ and $\sigma_{i,t}^2$ change.}
	\label{fig: F1}  \vspace{-0.3cm}
\end{figure}

In Figure \ref{fig: F1}, we reported the F1 score to measure the accuracy of learned causal graphs in both scenarios: one with only changing causal strengths (Figure \ref{fig: F1}(a)) and the other with changes in both $b_{ij,t}$ and $\sigma_{i,t}^2$ (Figure \ref{fig: F1}(b)). From the figure, one can see that our proposed method gives the best performance (the highest F1 score) in all cases, and the accuracy slightly increase along with sample size. The nonparametric method CD-NOD has the second-best performance. CD-NOD assumes that the changes are smooth, and in practice, it may be affected by inappropriately chosen kernel widths and significance level, and may need a large sample for good performance. The other three methods do not perform as well. 
IB and MC likely under-perform because they are designed for multi-domain systems and thus may not work well for the changes considered here. Similarly, LiNGAM is designed for fixed causal models and thus not appropriate for nonstationary data.



Then we did forecasting by making use of the estimated time-varying causal model. For each realization, we further simulated 10 values for the processes, and predicted the values of each process with one-step ahead prediction. We compared the proposed method with a collection of methods which do not consider the underlying causal model, including (vanilla) Lasso \cite{Lasso_Tibshirani96}, window-based Lasso, Kalman filtering (KF) \cite{Kalman60}, state-space model estimated with CPF-AS (denoted by SSM(CPF)), and Gaussian process (GP) regression \cite{Rasmussen06}. We used all the remaining processes as predictors for the target process. Particularly, the window size for window-based Lasso was 100.


\begin{figure}[htp!]
			\setlength{\abovecaptionskip}{15pt}
		\setlength{\belowcaptionskip}{15pt}
	\vspace{-.08cm}
		\includegraphics[width = 0.45\textwidth, height = 4cm, trim={0cm 0cm 1cm 0.2cm},clip]{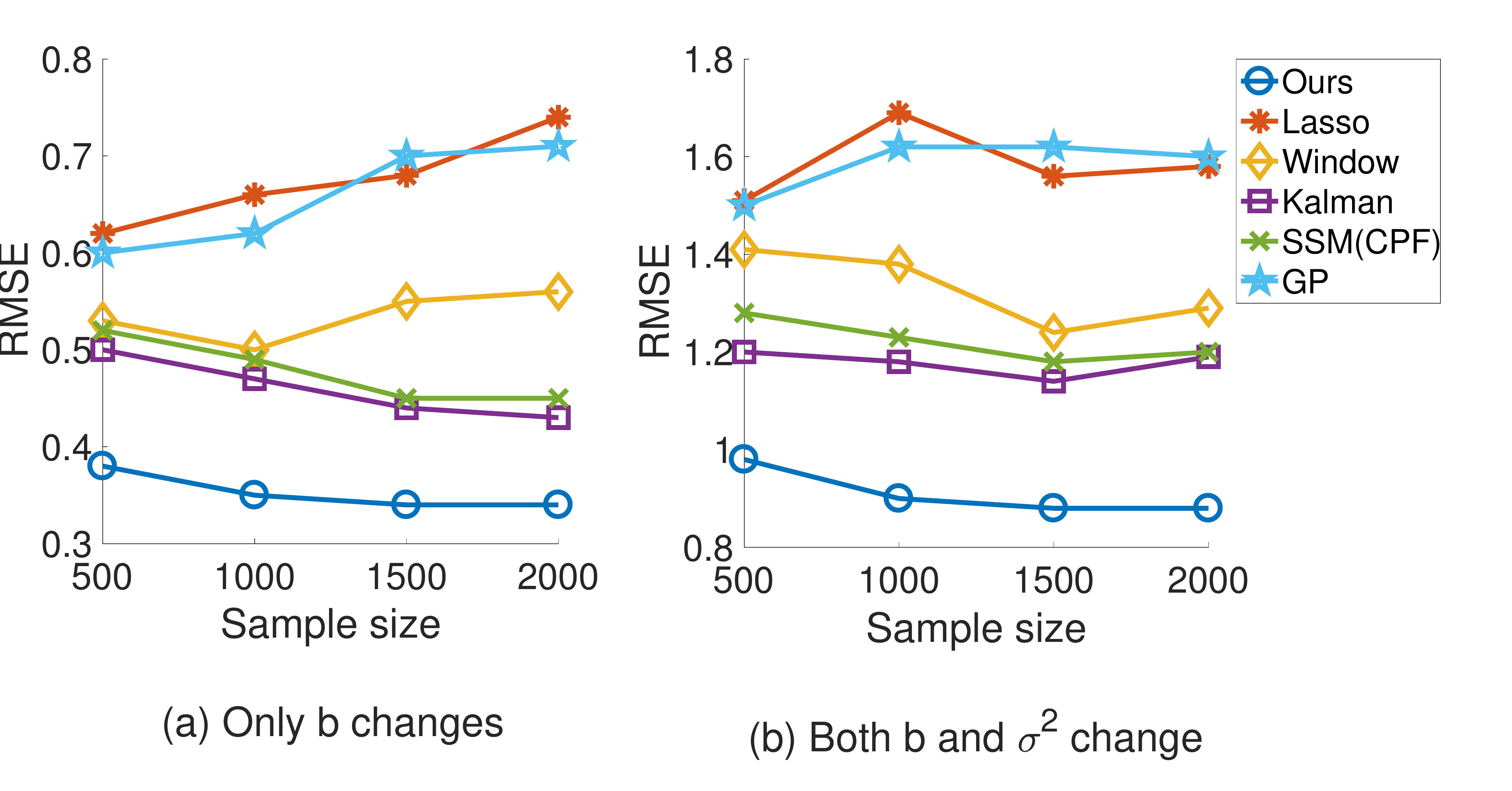} 
	\vspace{-0.2cm}
		\caption{RMSE of the forecasts when (a) only $b_{ij,t}$ change and when (b) both $b_{ij,t}$ and $\sigma_{i,t}^2$ change.} 
		\label{fig: RMSE} \vspace{-0.5cm}
	\end{figure}

\vspace{-0.1cm}
We calculated the root-mean-squared error (RMSE) of the predicted 10 values to evaluate the forecasting performance. We first did paired, one-sided Wilcoxon signed rank test between our method and each of the remaining ones  \cite{Nonpara_infer}, across the two settings (with constant and changing noise variances, respectively) and across the four different sample sizes. Our methods significantly outperform all others in all cases, with the highest $p$-value $0.018$ for the comparison with KF, $0.005$ with SSM(CPF), $ \times 10^{-4}$ with Lasso, $\times 10^{-3}$ with window-based Lasso, and $10^{-4}$with GP. For illustrative purposes, 
Figure \ref{fig: RMSE} shows the mean of RMSE across different processes and parameter settings; in (a), only causal strengths $b_{ij,t}$ change, and in (b) both $b_{ij,t}$ and noise variances $\sigma_{i,t}^2$ change. 
We can see that the RMSE generally decreases with sample size.  The Lasso and GP additionally do not consider the change of the model, and not surprisingly perform worse than others.

\begin{SCfigure}
		\vspace{-0.1cm}
	\includegraphics[width = 0.22\textwidth, trim={0cm 0.4cm 0cm 0.3cm},clip]{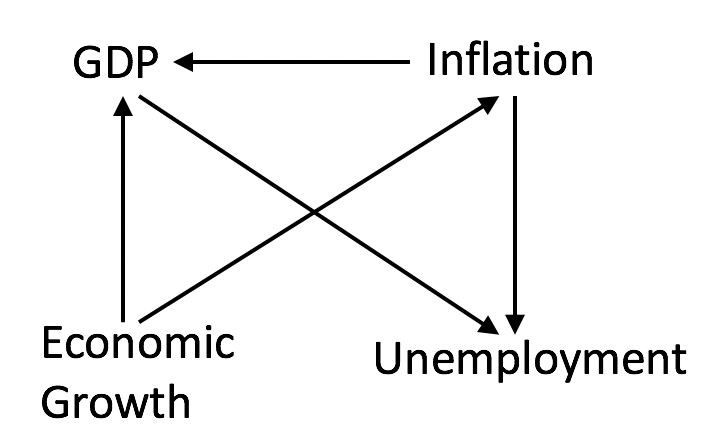} 
	\caption{Identified contemporaneous causal relationships between GDP, inflation, economic growth, and unemployment.}
	\label{fig: F2}
	\vspace{-0.39cm}
\end{SCfigure}

\vspace{-0.2cm}
\paragraph{Real-World Economic data}
We investigated the causal relationships between Gross Domestic Product (GDP), inflation, economic growth, and unemployment rate, with quarterly data from 1965 to 2017 in the USA \footnote{Downloaded from https://www.theglobaleconomy.com/.}. The data are normalized by subtracting the mean and dividing them by the standard deviation. We applied model (\ref{SSM}) to estimate contemporaneous causal relations between the four macroeconomic variables. From our model, we found that inflation and economic growth affect GDP, that economic growth influences inflation, and that unemployment is directly influenced by GDP and inflation; see Figure \ref{fig: F2}. These findings seem consistent with domain knowledge \footnote{For instance, see https://financialnerd.com/three-pillars-economy-inflation-gdp-unemployment/.}: for example, inflation increases the cost of products and leads to a decline in production, thus causing GDP to fall; 
economic growth  gradually increases the price level of all goods, thereby causing inflation; inflation may increase unemployment because of the decline in competitiveness and export demand. 

\vspace{-.45cm}
\begin{table}[htp]
	\caption{RMSE of the forecasts on inflation (2007 - 2017). }
	\begin{tabular}{lllll}
		\hline
		Methods          & RMSE &  & Methods      & RMSE \\ \hline
		Ours             & \textbf{0.32} &  & Lasso        & 0.38 \\ \hline
		Kalman filtering & 0.42 &  & Window Lasso & 0.37 \\ \hline
		SSM (CPF)        & 0.43 &  &    GP          &   0.37  \\ \hline
	\end{tabular}
	\label{Table: eco}
	\vspace{-.2cm}
\end{table}

We then forecasted inflation from  2007 to 2017 with one-step prediction.  We also included one-lagged time series as predictors. The RMSE on the normalized data is given in Table \ref{Table: eco}. Our method gives the best forecasting accuracy, as indicated by the lowest RMSE.

\vspace{-0.3cm}
\section{Conclusion}
In this paper, we formalized causal discovery and forecasting in nonstationary environments under the framework of nonlinear state-space models. We allowed changes in causal strengths, as well as noise variances. We showed that nonstationarity helps causal model identification, and that causal knowledge improves interpretability and forecasting accuracy. The proposed methods showed promising results on macroeconomic data. As future work, we will extend our methods to cover nonlinear causal relationships, to partially observable processes, as studied in \cite{hidden_series_ICML15}, and to causal models with instantaneous cycles.

\section*{Acknowledgements} 
We thank Jeff Adams for helping to revise the paper. We would like to acknowledge the supported by National Institutes of Health under Contract No. NIH-1R01EB022858-01, FAINR01EB022858, NIH-1R01LM012087, NIH-5U54HG008540-02, and FAIN- U54HG008540, by the United States Air Force under Contract No. FA8650-17-C-7715, and by National Science Foundation EAGER Grant No. IIS-1829681. The National Institutes of Health, the U.S. Air Force, and the National Science Foundation are not responsible for the views reported in this article.

\onecolumn
\icmltitle{Supplementary Material for ``Causal Discovery and Forecasts in Nonstationary Environments with State-Space Models''}

\newcommand{\beginsupplement}{%
	\setcounter{table}{0}
	\renewcommand{\thetable}{S\arabic{table}}%
	\setcounter{figure}{0}
	\renewcommand{\thefigure}{S\arabic{figure}}%
	\setcounter{algorithm}{0}
	\renewcommand{\thealgorithm}{S\arabic{algorithm}}%
	\setcounter{section}{0}
	\renewcommand{\thesection}{S\arabic{section}}%
	\setcounter{theorem}{0}
	\renewcommand{\thetheorem}{S\arabic{theorem}}%
	\setcounter{lemma}{0}
	\renewcommand{\thelemma}{S\arabic{lemma}}%
}

	\beginsupplement

	\section{Proof of Theorem 1}
	
	%

	Before the proof, we first give Lemma \ref{Lemma: S1}, which gives the identifiability of parameters in varying coefficients regression models. This result will be used in the proof of Theorem 1.
	\begin{lemma} [\cite{Identifiability_SSM}]
		The varying coefficients regression model takes on the following form: 
		\begin{equation}
		\left \{
		\begin{array}{llll} 
		y_t  =  \sum_i b_{i,t} x_{i,t} + e_t, \\
		b_{i,t}  =  \alpha_{i,0} + \sum_p \alpha_{i,p} b_{i,t-p} + \epsilon_{i,t},
		\end{array}\right.
		\end{equation}
		where $y_t$ is the scalar valued dependent variable and $x_{i,t}$ is the independent variable which we have observations. The additive error, $e_t$, represents a stationary zero mean white noise process, i.e.,  $E[e_t] = 0$ and $E[e_t e_{t'}] = \sigma_e^2 \delta_{tt'}$, $E[\epsilon_{i,t}] = 0$, where $\sigma_e^2 < \infty$ and $\delta_{tt'}$ is the delta function. Similarly,  $E[\epsilon_{i,t}] = 0$ and $E[\epsilon_{i,t} \epsilon_{i,t'}] = \sigma_{\epsilon_i}^2 \delta_{tt'}$, for $\forall i \in \mathbf{N}^+$. 
		
		Then the parameters $\sigma_e^2$,  $\alpha_{i,0}$, $\alpha_{i,p}$, $\sigma_{\epsilon_i}^2$, for $\forall i,p \in \mathbf{N}^+$ are globally identifiable.
		\label{Lemma: S1}
	\end{lemma}
	
	Now we start to prove Theorem 1.
	\begin{proof}[Proof of Theorem 1]
		The proof of Theorem 1 contains two phases. In the first phase, we identify the root variable and corresponding causal parameters. In the second phase, we identify the remaining causal graph and corresponding parameters in a recursive way.
		
		\paragraph{Phase I}	Let $A_t = (I-B_t)^{-1}$. We define the following metric to characterize kurtosis of observed variables:
		\begin{equation}
		\begin{array}{ll}
		& E \big[ x_{i,t} x_{j,t} x_{k,t+p}  x_{l,t+p} \big] \\[2pt]
		=  &E \big[ \sum_a A_{ia,t}E_{a,t}  \sum_b A_{jb,t}E_{b,t}  \sum_c A_{kc,t+p}E_{c,t+p}  \sum_d A_{ld,t+p}E_{d,t+p} \big] \\[2pt]
		= &E \big[ \sum_a \sum_b \sum_c \sum_d A_{ia,t} A_{jb,t} A_{kc,t+p} A_{ld,t+p} E_{a,t}  E_{b,t}  E_{c,t+p}  E_{d,t+p} \big] \\ [2pt]
		= &\sum_a \sum_c E \big[ A_{ia,t} A_{ja,t} A_{kc,t+p} A_{lc,t+p} E_{a,t}  E_{a,t}  E_{c,t+p}  E_{c,t+p} \big] \\[2pt]
		= &\sum_a \sum_c \sigma_a^2 \sigma_c^2  E \big[ A_{ia,t} A_{ja,t} A_{kc,t+p} A_{lc,t+p} \big],   
		\end{array}
		\end{equation}
		where the third equation holds because only when $a=b$ and $c=d$, the expectation is not zero.
		
		By considering all combinations of $i, j, k, \text{and } l$, we can organize the above kurtosis in the matrix form,
		\begin{equation}
		\begin{array}{ll}
		&\mathbf{S}(t,t+p)\\[2pt]
		= & E \big[ (A_t \Sigma_E A_t^T) \otimes (A_{t+p} \Sigma_E A_{t+p}^T)  \big] \\[2pt]
		= & E \big[ (I-B_t)^{-1} \Sigma_E (I-B_t)^{-T}) \otimes ((I-B_{t+p})^{-1} \Sigma_E (I-B_{t+p})^{-T})  \big].
		\end{array}
		\end{equation}
		where $\mathbf{S}(t,t+p)$ is a $n^2 \times n^2$ matrix, $\otimes$ denotes Kronecker product, and 
		\begin{equation*}
		\begin{array}{ll}
		& E \big[ x_{i,t} x_{j,t} x_{k,t+p}  x_{l,t+p} \big] \\ [2pt]
		& = \mathbf{S}(t,t+p)_{ijkl} \\ [2pt]
		& = E \big[ \{ (I-B_t)^{-1} \Sigma_E (I-B_t)^{-T})\}_{ij} \cdot \{(I-B_{t+p})^{-1} \Sigma_E (I-B_{t+p})^{-T})\}_{kl}  \big].
		\end{array}
		\end{equation*}
		
		If the underlying graph is a DAG, then $(I-B_t)^{-1}$ can be reformulated as
		\begin{equation*}
		(I-B_t)^{-1} = \sum_{r = 0}^{m-1} B_t^r,
		\end{equation*}
		and thus
		\begin{equation}
		\mathbf{S}(t,t+p) = E \big[ \big ( (\sum_{r = 0}^{m-1} B_t^r) \Sigma_E  (\sum_{r = 0}^{m-1} B_t^{T r}) \big) \otimes \big( (\sum_{r = 0}^{m-1} B_{t+p}^r) \Sigma_E  (\sum_{r = 0}^{m-1} B_{t+p}^{T r}) \big) \big].
		\end{equation}
		
		Particularly, let us consider the case when $i=j=k=l$ and for notation simplicity, denote  $S(t,t+p)_{ijkl}$ by $S(t,t+p)_{i}$. Then
		\begin{equation}
		S(t,t+p)_i = E[x_{i,t}^2  x_{i,t+p}^2].
		\end{equation}
		Let $r_0$ be the index of the root cause, and $\mathbf{V}_s = \mathbf{V} \backslash r_0$ denote the indices of the remaining processes, with $ \mathbf{V} = \{1,\cdots,m\}$. Then we will have
		\begin{equation}
		\begin{array}{ll}
		S(t,t+p)_{r_0} - S(t,t)_{r_0} = 0,\\ [2pt]
		S(t,t+p)_{r_s} - S(t,t)_{r_s} < 0, \quad \forall r_s \in \mathbf{V}_s,
		\end{array}
		\label{root}
		\end{equation}
		for any $p \in \mathbf{N}^+$. The reason is that the root cause does not receive changing influences from other processes.
		
		Let us now give the detailed proof procedure of (\ref{root}). 
		Suppose that we have known the causal order, denoted by $\pi$. Let us first see a few examples of the concrete representations of $S(t,t+p)_i $. 
		\begin{enumerate}
			\item  For the root cause $\pi(1)$, it is easy to get $S(t,t+p)_{\pi(1)}  = (\sigma_{\pi(1)}^2)^2$, which is irrelevant to $p$.
			Thus, $$S(t,t+p)_{\pi(1)} - S(t,t)_{\pi_{(1)}} = 0.$$
			
			\item For $\pi(2)$, we have
			\begin{equation*} 
			\begin{array}{ll}
			& S(t,t+p)_{\pi(2)} \\ [2pt]
			= & (\sigma_{\pi(2)}^2)^2 + \sigma_{\pi(1)}^2 \sigma_{\pi(2)}^2 E[b_{{\pi(2)}{\pi(1)},t+p}^2] + \sigma_{\pi(1)}^2 \sigma_{\pi(2)}^2 E[b_{{\pi(2)}{\pi(1)},t}^2]  + (\sigma_{\pi(1)}^2)^2 E[b_{{\pi(2)}{\pi(1)},t}^2 b_{{\pi(2)}{\pi(1)},t+p}^2].
			\end{array}
			\end{equation*}
			
			Thus, 
			\begin{equation*}
			\begin{array}{ll}
			& S(t,t+p)_{\pi(2)} - S(t,t)_{\pi(2)}  \\ [2pt]
			= & (\sigma_{\pi(1)}^2)^2 \bigg (E[b_{{\pi(2)}{\pi(1)},t}^2 b_{{\pi(2)}{\pi(1)},t+p}^2] - E[b_{{\pi(2)}{\pi(1)},t}^2 b_{{\pi(2)}{\pi(1)},t}^2] \bigg) \\ [2.5pt]
			= & (\sigma_{\pi(1)}^2)^2 \bigg (E[b_{{\pi(2)}{\pi(1)},t}^2 \cdot (\alpha_{\pi(2)\pi(1)}^p b_{_{\pi(2)\pi(1)},t} + \alpha_{\pi(2)\pi(1)}^{p-1} \epsilon_{t+1} +\cdots + \alpha_{\pi(2)\pi(1)}^{0} \epsilon_{t+p})^2] - E[b_{{\pi(2)}{\pi(1)},t}^2 b_{{\pi(2)}{\pi(1)},t}^2] \bigg)\\ [2.5pt]
			= & (\sigma_{\pi(1)}^2)^2  \cdot 2 (\alpha_{\pi(2)\pi(1)}^{2p}-1) \frac{(w_{\pi(2)\pi(1)})^2 }{(1-\alpha_{\pi(2)\pi(1)}^2)^2} \\[2pt]
			< & 0,
			\end{array}
			\end{equation*}
			where $w_{\pi(2)\pi(1)}$ is the noise variance in the autoregressive model of $b_{\pi(2)\pi(1),t}$. $S(t,t+p)_{\pi(2)} - S(t,t)_{\pi(2)} <0$ always holds because $\alpha_{\pi(2)\pi(1)}^{2p}-1<0, \forall p \in \mathbf{N}^+$, since $\alpha_{\pi(2)\pi(1)} \in (-1,1)$.
			
			\item For $\pi(3)$, we have
			\begin{equation*}
			\begin{array}{ll}
			& S(t,t+p)_{\pi(3)} \\ [2pt]
			
			= & (\sigma_{\pi(1)}^2)^2 E[b_{{\pi(3)}{\pi(1)},t}^2 b_{{\pi(3)}{\pi(1)},t+p}^2] \\ [2pt]
			& + 4 (\sigma_{\pi(1)}^2)^2 E[b_{{\pi(2)}{\pi(1)},t} b_{{\pi(2)}{\pi(1)},t+p}] E[b_{{\pi(3)}{\pi(1)},t} b_{{\pi(3)}{\pi(1)},t+p}] E[b_{{\pi(3)}{\pi(1)},t} b_{{\pi(3)}{\pi(2)},t+p}] \\ [2pt]
			
			& + (\sigma_{\pi(1)}^2)^2 E[b_{{\pi(2)}{\pi(1)},t}^2 b_{{\pi(2)}{\pi(1)},t+p}^2] E[b_{{\pi(3)}{\pi(2)},t}^2 b_{{\pi(3)}{\pi(2)},t+p}^2] + (\sigma_{\pi(2)}^2)^2 E[b_{{\pi(3)}{\pi(2)},t}^2 b_{{\pi(3)}{\pi(2)},t+p}^2] \\[2pt]
			& + (\sigma_{\pi(3)}^2)^2 + (\sigma_{\pi(1)}^2)^2 E[b_{{\pi(2)}{\pi(1)},t+p}^2] E[b_{{\pi(3)}{\pi(1)},t+p}^2] E[b_{{\pi(3)}{\pi(2)},t+p}^2] \\[2pt]
			& + \sigma_{\pi(1)}^2 \sigma_{\pi(2)}^2 E[b_{{\pi(3)}{\pi(1)},t}^2]E[b_{{\pi(3)}{\pi(2)},t+p}^2] + \sigma_{\pi(1)}^2 \sigma_{\pi(3)}^2 E[b_{{\pi(3)}{\pi(1)},t}^2]\\[2pt]
			
			&  + (\sigma_{\pi(1)}^2)^2 E[b_{{\pi(2)}{\pi(1)},t}^2] E[b_{{\pi(3)}{\pi(1)},t+p}^2] E[b_{{\pi(3)}{\pi(2)},t}^2] + \sigma_{\pi(1)}^2 \sigma_{\pi(2)}^2 E[b_{{\pi(3)}{\pi(1)},t}^2]E[b_{{\pi(3)}{\pi(2)},t+p}^2] \\[2pt]
			& + \sigma_{\pi(1)}^2 \sigma_{\pi(2)}^2 E[b_{{\pi(3)}{\pi(1)},t}^2]E[b_{{\pi(3)}{\pi(2)},t+p}^2]\\[2pt]		         	
			
			& + \sigma_{\pi(1)}^2 \sigma_{\pi(2)}^2 E[b_{{\pi(3)}{\pi(1)},t+p}^2] E[b_{{\pi(3)}{\pi(2)},t}^2] + \sigma_{\pi(1)}^2 \sigma_{\pi(2)}^2 E[b_{{\pi(2)}{\pi(1)},t+p}^2] E[b_{{\pi(3)}{\pi(2)},t+p}^2] \\[2pt]
			&+ \sigma_{\pi(2)}^2 \sigma_{\pi(3)}^2 E[b_{{\pi(3)}{\pi(2)},t}^2] + \sigma_{\pi(1)}^2 \sigma_{\pi(3)}^2 E[b_{{\pi(3)}{\pi(1)},t+p}^2] \\[2pt]
			&+ \sigma_{\pi(1)}^2 \sigma_{\pi(3)}^2 E[b_{{\pi(2)}{\pi(1)},t+p}^2] E[b_{{\pi(3)}{\pi(2)},t+p}^2] + \sigma_{\pi(2)}^2 \sigma_{\pi(3)}^2 E[b_{{\pi(3)}{\pi(2)},t+p}^2].
			\end{array}
			\end{equation*}
			
			Thus,
			\begin{equation*}
			\begin{array}{lll}
			& S(t,t+p)_{\pi(3)} - S(t,t)_{\pi(3)} \\ [2pt]
			
			= & (\sigma_{\pi(1)}^2)^2 \bigg ( E[b_{{\pi(3)}{\pi(1)},t}^2 b_{{\pi(3)}{\pi(1)},t+p}^2] - E[b_{{\pi(3)}{\pi(1)},t}^2 b_{{\pi(3)}{\pi(1)},t}^2]  \bigg) \\[2pt]
			
			& + 4 (\sigma_{\pi(1)}^2)^2 \bigg ( E[b_{{\pi(2)}{\pi(1)},t} b_{{\pi(2)}{\pi(1)},t+p}] E[b_{{\pi(3)}{\pi(1)},t} b_{{\pi(3)}{\pi(1)},t+p}] E[b_{{\pi(3)}{\pi(2)},t} b_{{\pi(3)}{\pi(2)},t+p}] \\[2pt]
			& \qquad \qquad \qquad - E[b_{{\pi(2)}{\pi(1)},t} b_{{\pi(2)}{\pi(1)},t}] E[b_{{\pi(3)}{\pi(1)},t} b_{{\pi(3)}{\pi(1)},t}] E[b_{{\pi(3)}{\pi(2)},t} b_{{\pi(3)}{\pi(2)},t}]\bigg )\\[2pt]
			
			& + (\sigma_{\pi(1)}^2)^2 \bigg ( E[b_{{\pi(2)}{\pi(1)},t}^2 b_{{\pi(2)}{\pi(1)},t+p}^2] E[b_{{\pi(3)}{\pi(2)},t}^2 b_{{\pi(3)}{\pi(2)},t+p}^2] - E[b_{{\pi(2)}{\pi(1)},t}^2 b_{{\pi(2)}{\pi(1)},t}^2] E[b_{{\pi(3)}{\pi(2)},t}^2 b_{{\pi(3)}{\pi(2)},t}^2] \bigg)\\[2pt]
			
			& + (\sigma_{\pi(2)}^2)^2 \bigg ( E[b_{{\pi(3)}{\pi(2)},t}^2 b_{{\pi(3)}{\pi(2)},t+p}^2] - E[b_{{\pi(3)}{\pi(2)},t}^2 b_{32,t}^2] \bigg) \\[2pt]
			
			&  +  \sigma_{\pi(1)}^2 \sigma_{\pi(2)}^2 \bigg ( E[b_{{\pi(3)}{\pi(1)},t}^2]E[b_{{\pi(3)}{\pi(2)},t+p}^2] - E[b_{{\pi(3)}{\pi(1)},t}^2]E[b_{{\pi(3)}{\pi(2)},t}^2] \bigg ) \\[2pt]       	
			
			& + \sigma_{\pi(1)}^2 \sigma_{\pi(1)}^2 \bigg ( E[b_{{\pi(2)}{\pi(1)},t+p}^2] E[b_{{\pi(3)}{\pi(2)},t+p}^2] - E[b_{{\pi(2)}{\pi(1)},t}^2] E[b_{{\pi(3)}{\pi(2)},t}^2] \bigg) \\[2pt]
			
			< & 0 			     
			\end{array}
			\end{equation*}
			
		\end{enumerate}

		Generally, we have the following form for variable with the $n$th $(n>1)$ order:
		\begin{equation}
		\begin{array}{ll}
		& S(t,t+p)_{\pi(n)} \\[2pt]
		= & \bigg( \sum_{i=1}^{\pi(n)-1} (\sum_{k>1,l} p_{\pi(n) i,t}^{(k,l)})^2 \sigma_i^2 + \sigma_{\pi(n)}^2 \bigg) \cdot \bigg( \sum_{i'=1}^{\pi(n)-1} (\sum_{k'>1,l'} p_{\pi(n) i',t+p}^{(k',l')})^2 \sigma_{i'}^2 + \sigma_{\pi(n)}^2 \bigg),
		\end{array}
		\label{Snn}			      
		\end{equation}
		where $p_{\pi(n) i,t}^{(k,l)}$ indicates the multiplication of causal coefficients of the $l$th directed path \footnote{A directed path is defined as a sequence of edges (or arcs) which connect a sequence of vertices, and the edges are all directed in the same direction.} from node $i$ to node $\pi(n)$ with path length $k$. 
		For example, for path with length 1, $p_{\pi(n) i,t}^{(1)} = b_{\pi(n) i,t}$; for path with length 2, $p_{\pi(n) i,t}^{(2)} = b_{\pi(n) j,t}b_{ji,t}$, with $i<j<\pi(n)$; for path with length 3, $p_{\pi(n) i,t}^{(3)} = b_{\pi(n) j_1,t}b_{j_1j_2,t}b_{j_2i,t}$, with $i<j_2<j_1<\pi(n)$. The longest path length from node $i$ to node $\pi(n)$ is $\pi(n)-i$.
		
		By expanding (\ref{Snn}), we have
		\begin{equation*}
		\begin{array}{ll}
		& S(t,t+p)_{\pi(n)} \\[2pt]
		= & (\sigma_{\pi(n)}^2)^2  + \sum_{i<\pi(n)} E[ b_{\pi(n) i,t}^2  b_{\pi(n) i,t+p}^2] (\sigma_{i}^2)^2\\[2pt]
		& + \sum_{i<\mathbf{j}<\pi(n)} E[ b_{\pi(n) \mathbf{j},t}^2  b_{\pi(n) \mathbf{j},t+p}^2] E[ b_{\mathbf{j} i,t}^2  b_{\mathbf{j} i,t+p}^2] (\sigma_{i}^2)^2 \\[2pt]
		& + \sum_{i<\mathbf{j},\mathbf{j'}<\pi(n),\mathbf{j} \neq \mathbf{j'} } 4 E[ b_{\pi(n) \mathbf{j},t}  b_{\pi(n) \mathbf{j},t+p}] E[b_{\mathbf{j} i,t}  b_{\mathbf{j} i,t+p}]  E[b_{\pi(n)\mathbf{j'},t}  b_{\pi(n) \mathbf{j'},t+p}]  E[b_{\mathbf{j'} i,t}  b_{\mathbf{j'} i,t+p}] (\sigma_{i}^2)^2  \\[2pt]
		& + \sum_{i<\mathbf{j}<\pi(n),i'<\mathbf{j'}<\pi(n), \mathbf{j} \neq \mathbf{j'} \cup i \neq i'} E[b_{\pi(n) \mathbf{j},t}^2 b_{\mathbf{j}i,t}^2 b_{\pi(n) \mathbf{j'},t+p}^2 b_{\mathbf{j'}i',t+p}^2] \sigma_{i}^2 \sigma_{i'}^2 \\[2pt]
		\end{array}
		\label{Snn2}			      
		\end{equation*}
		where $\mathbf{j}$ and $\mathbf{j'}$ can be a series of indices, $\mathbf{j} = (j_1, j_2, \cdots, j_k)$ with $\pi(n) > j_1 > j_2> \cdots i$, and $\mathbf{j'} = (j'_1, j'_2, \cdots, j'_k)$ with $\pi(n) > j'_1 > j'_2> \cdots i$, and 
		\begin{equation*}
		\begin{array}{ll}
		&  E[ b_{\pi(n) \mathbf{j},t}^2  b_{\pi(n) \mathbf{j},t+p}^2] E[ b_{\mathbf{j} i,t}^2  b_{\mathbf{j} i,t+p}^2] 
		= E[ b_{\pi(n) j_1,t}^2  b_{\pi(n) j_1,t+p}^2] E[ b_{j_1 j_2,t}^2  b_{j_1 j_2,t+p}^2] \cdots E[ b_{j_k i,t}^2  b_{j_k i,t+p}^2], \\[8pt]
		& E[ b_{\pi(n) \mathbf{j},t}  b_{\pi(n) \mathbf{j},t+p}] E[b_{\mathbf{j} i,t}  b_{\mathbf{j} i,t+p}] 
		= E[ b_{\pi(n) j_1,t}  b_{\pi(n) j_1,t+p}] E[ b_{j_1 j_2,t}  b_{j_1 j_2,t+p}] \cdots E[b_{j_k i,t}  b_{j_k i,t+p}], \\[8pt]
		& E[ b_{\pi(n) \mathbf{j'},t}  b_{\pi(n) \mathbf{j'},t+p}] E[b_{\mathbf{j'} i,t}  b_{\mathbf{j'} i,t+p}] 
		=  E[ b_{\pi(n) j'_1,t}  b_{\pi(n) j'_1,t+p}] E[ b_{j'_1 j'_2,t}  b_{j'_1 j'_2,t+p}] \cdots E[b_{j'_k i,t}  b_{j'_k i,t+p}],\\[8pt]
		& E[b_{\pi(n)\mathbf{j},t}^2 b_{\mathbf{j}i,t}^2 b_{\pi(n)\mathbf{j'},t+p}^2 b_{\mathbf{j'}i',t+p}^2] 
		=  E[b_{\pi(n) j_1,t}^2 b_{j_1 j_2,t}^2 \cdots b_{j_k i,t}^2 b_{n j'_1,t+p}^2 b_{j'_1 j'_2,t+p}^2 \cdots b_{j'_k i',t+p}^2].	
		\end{array}
		\end{equation*}
		
		
		Then 
		\begin{equation*}
		\begin{array}{ll}
		& S(t,t+p)_{\pi(n)} - S(t,t)_{\pi(n)}\\[2pt]
		= &\sum_{i<\pi(n)} \bigg( E[ b_{\pi(n) i,t}^2  b_{\pi(n) i,t+p}^2] - E[ b_{\pi(n) i,t}^2  b_{\pi(n) i,t}^2] \bigg ) (\sigma_{i}^2)^2\\
		& + \sum_{i<\mathbf{j}<\pi(n)} \bigg( E[ b_{\pi(n) \mathbf{j},t}^2  b_{\pi(n) \mathbf{j},t+p}^2] E[ b_{\mathbf{j} i,t}^2  b_{\mathbf{j} i,t+p}^2] - E[ b_{\pi(n) \mathbf{j},t}^2  b_{\pi(n) \mathbf{j},t}^2] E[ b_{\mathbf{j} i,t}^2  b_{\mathbf{j} i,t}^2]  \bigg) (\sigma_{i}^2)^2 \\
		& + \sum_{i<\mathbf{j},\mathbf{j'}<\pi(n), \mathbf{j} \neq \mathbf{j'}} 4 \bigg( E[ b_{\pi(n) \mathbf{j},t}  b_{\pi(n) \mathbf{j},t+p}] E[b_{\mathbf{j} i,t}  b_{\mathbf{j} i,t+p}]  E[b_{\pi(n) \mathbf{j'},t}  b_{\pi(n) \mathbf{j'},t+p}]  E[b_{\mathbf{j'} i,t}  b_{\mathbf{j'} i,t+p}] \\
		& \qquad \qquad \qquad \qquad \qquad - E[ b_{\pi(n)\mathbf{j},t}^2] E[b_{\mathbf{j} i,t}^2]  E[b_{\pi(n) \mathbf{j'},t}^2]  E[b_{\mathbf{j'} i,t}^2  ] \bigg) (\sigma_{i}^2)^2\\
		& + \sum_{i<\mathbf{j}<\pi(n),i'<\mathbf{j'}<\pi(n), \mathbf{j} \neq \mathbf{j'} \cup i \neq i'} \bigg( E[b_{\pi(n) \mathbf{j},t}^2 b_{\mathbf{j}i,t}^2 b_{n\mathbf{j'},t+p}^2 b_{\mathbf{j'}i,t+p}^2] - E[b_{\pi(n) \mathbf{j},t}^2 b_{\mathbf{j}i,t}^2 b_{\pi(n) \mathbf{j'},t}^2 b_{\mathbf{j'}i,t}^2] \bigg) \sigma_{i}^2 \sigma_{i'}^2,
		\end{array}
		\label{Snn3}			      
		\end{equation*}
		where
		\begin{equation*}
		\begin{array}{ll}
		E[ b_{\pi(n) i,t}^2  b_{\pi(n) i,t+p}^2] - E[ b_{\pi(n) i,t}^2  b_{\pi(n) i,t}^2]  = 2 (\alpha_{\pi(n) i}^{2p}-1) \frac{(w_{\pi(n) i})^2 }{(1-\alpha_{\pi(n) i}^2)^2} 
		< 0, \\
		~\\
		E[ b_{\pi(n) \mathbf{j},t}^2  b_{\pi(n)\mathbf{j},t+p}^2] E[ b_{\mathbf{j} i,t}^2  b_{\mathbf{j} i,t+p}^2] - E[ b_{\pi(n) \mathbf{j},t}^2  b_{\pi(n) \mathbf{j},t}^2] E[ b_{\mathbf{j} i,t}^2  b_{\mathbf{j} i,t}^2]  \\[5pt]
		= \big ( (1+2\alpha_{\pi(n) j}^{2p})(1+2\alpha_{ji}^{2p}) - 9 \big) \cdot \frac{(w_{\pi(n) j})^2 }{(1-\alpha_{\pi(n) j}^2)^2} \frac{(w_{ji})^2 }{(1-\alpha_{ji}^2)^2}<0,\\
		~\\
		E[ b_{\pi(n) \mathbf{j},t}  b_{\pi(n) \mathbf{j},t+p}] E[b_{\mathbf{j} i,t}  b_{\mathbf{j} i,t+p}]  E[b_{\pi(n) \mathbf{j'},t}  b_{\pi(n) \mathbf{j'},t+p}]  E[b_{\mathbf{j'} i',t}  b_{\mathbf{j'} i',t+p}] - E[ b_{\pi(n) \mathbf{j},t}^2] E[b_{\mathbf{j} i,t}^2]  E[b_{\pi(n) \mathbf{j'},t}^2]  E[b_{\mathbf{j'} i',t}^2  ]  \\[5pt]
		=   (\alpha_{\pi(n) j}^p \alpha_{\mathbf{j}i}^p \alpha_{\pi(n)\mathbf{j'}}^p \alpha_{\mathbf{j'} i'}^p - 1) E[b_{\pi(n) \mathbf{j},t}^2] E[b_{\mathbf{j} i,t}^2]  E[b_{\pi(n) \mathbf{j'},t}^2] E[b_{\mathbf{j'} i',t}^2] <0, \\
		~\\
		E[b_{n\mathbf{j},t}^2 b_{\mathbf{j}i,t}^2 b_{n\mathbf{j'},t+p}^2 b_{\mathbf{j'}i,t+p}^2] - E[b_{n\mathbf{j},t}^2 b_{\mathbf{j}i',t}^2 b_{n\mathbf{j'},t}^2 b_{\mathbf{j'}i,t}^2] \leq 0.
		\end{array}			     
		\end{equation*}	
		Thus, 
		\begin{equation}
		S(t,t+p)_{\pi(n)}  - S(t,t)_{\pi(n)} <0 \text{ for } n>1.
		\end{equation}

		Hence, $S(t,t+p)_{\pi(n)}  - S(t,t)_{\pi(n)} <0$ for $n>1$, and  $S(t,t+p)_{\pi(1)}  - S(t,t)_{\pi(1)}= 0$. Therefore, we can identify the root cause. Since $S(t,t+p)_{\pi(1)} = \sigma_{\pi(1)}^4$, the corresponding parameter $\sigma_{\pi(1)}^2$ is also identifiable.

		\paragraph{Phase II} After identifying the root cause, in the second phase, we then identify the causal model over the remaining variables in a recursive way.
		
		Suppose that we have known the causal order $\pi$. Then for each $x_{\pi(i)}$, its generating process can be reformulated as:
		\begin{equation}
		\begin{array}{ll}
		x_{\pi(i),t} = b^T_{\pi(i),t} x_{\pi_{1,\cdots,i-1},t} + e_{\pi(i),t} , \\
		b_{\pi(i),t}  = A_{\pi(i)}  b_{\pi(i),t-1}  + \epsilon_{\pi(i),t},
		\end{array}
		\label{DLM}
		\end{equation}
		where $x_{\pi_{1,\cdots,i-1},t} = [x_{\pi(1),t}, \cdots, x_{\pi(i-1),t}]$, representing potential causes to $x_{\pi(i),t}$. 
		
		Lemma \ref{Lemma: S1} has shown that parameters of the varying coefficients regression model are identifiable. Thus,  parameters in (\ref{DLM}) are identifiable; that is, the corresponding parameters $A_{\pi(i)}$, $\sigma_{e_{\pi(i)}}^2$(the variance of $e_{\pi(i),t}$), and $\sigma_{\epsilon_{\pi(i)}}^2$(the variance of $\epsilon_{\pi(i),t}$) are all identifiable, given the causal order.

		
		We define a node's \textit{level} in a acyclic graph the number of nodes in the directed path from the root to the node. For instance, the root has level 1, and any one of its adjacent nodes has level 2.
		
		Suppose that we have identified the causal model of variables at the first $n$ levels. Next, we will identify variables at the $(n+1)$th level and their corresponding parameters. Let $\mathbf{V}_n$ represent variable indices of the first $n$ processes, and let $\bar{\mathbf{V}}_{n} = \mathbf{V} \backslash \mathbf{V}_n$. 
		In the following, we will show that for node $r_{s} \in \bar{\mathbf{V}}_{n}$ which is at the $(n+1)$th level, $S(t,t+p)_{r_s} $ can be totally explained by a linear combination
		of cross-statistics of different orders of $x_{\mathbf{V}_n,t}$, but not for other nodes. 
		
		We denote $r_{s} \in \bar{\mathbf{V}}_{n}$ at the $(n+1)$th level as $\pi(n+1)$. Then we have
		\begin{equation}
		\begin{array}{ll}
		& S(t,t+p)_{\pi(n+1)} \\ [5pt]
		= & E \big [x_{\pi(n+1),t}^2 x_{\pi(n+1),t+p}^2 \big] \\ [5pt]
		= & E \big[ ( b^T_{\pi(n+1),t} x_{\mathbf{V}_n,t} + e_{\pi(n+1),t} )^2  ( b^T_{\pi(n+1),t+p} x_{\mathbf{V}_n,t+p} + e_{\pi(n+1),t+p} )^2 \big] \\[5pt]
		= & E[ (b^T_{\pi(n+1),t} x_{\mathbf{V}_n,t})^2 (b^T_{\pi(n+1),t+p} x_{\mathbf{V}_n,t+p})^2] + E[ e_{\pi(n+1)^2,t}] E[ e_{\pi(n+1)^2,t+p} ] \\ [2pt]
		&+ E[ (b^T_{\pi(n+1),t} x_{\mathbf{V}_n,t})^2] E[ e_{\pi(n+1),t+p}^2 ] + E[ (b^T_{\pi(n+1),t+p} x_{\mathbf{V}_n,t+p})^2] E[ e_{\pi(n+1),t}^2 ] \\[5pt]
		= & E[ (b^T_{\pi(n+1),t} x_{\mathbf{V}_n,t})^2 (b^T_{\pi(n+1),t+p} x_{\mathbf{V}_n,t+p})^2] + (\sigma_{e_{\pi(n+1)}}^2)^2 + E[ (b^T_{\pi(n+1),t} x_{\mathbf{V}_n,t})^2] \sigma_{e_{\pi(n+1)}}^2 \\ [2pt]
		&+ E[ (b^T_{\pi(n+1),t+p} x_{\mathbf{V}_n,t+p})^2] \sigma_{e_{\pi(n+1)}}^2 \\[5pt]
		
		=&  \sum_{i,j} E[ b_{i,t}^2 b_{j,t+p}^2 ] E [ x_{i,t}^2 x_{j,t+p}^2 ] + 2 \sum_{i,j \neq k} E[ b_{i,t}^2 b_{j,t+p} b_{k,t+p} ] E [ x_{i,t}^2 x_{j,t+p} x_{k,t+p} ] + (\sigma_{e_{\pi(n+1)}}^2)^2  \\[2pt]
		&+ 2 \sum_{i,j \neq k} E[ b_{i,t+p}^2 b_{j,t} b_{k,t} ] E [ x_{i,t+p}^2 x_{j,t} x_{k,t} ] 
		+ 4 \sum_{i \neq j, k \neq l} E[ b_{i,t} b_{j,t} b_{k,t+p} b_{l,t+p}] E [  x_{i,t} x_{j,t} x_{k,t+p} x_{l,t+p}] \\[2pt]
		& + \sum_{i} E[ b_{i,t}^2] E[ x_{i,t}^2] \sigma_{e_{\pi(n+1)}}^2 + \sum_{i \neq j} 2 E[ b_{i,t} b_{j,t}] E[ x_{i,t} x_{j,t}] \sigma_{e_{\pi(n+1)}}^2   + \sum_{i} E[ b_{i,t+p}^2] E[ x_{i,t+p}^2] \sigma_{e_{\pi(n+1)}}^2 \\[2pt]
		&+ \sum_{i \neq j} 2 E[ b_{i,t+p} b_{j,t+p}] E[ x_{i,t+p} x_{j,t+p}] \sigma_{e_{\pi(n+1)}}^2 \\[5pt]
		
		= & \sum_{i \neq j} E[ b_{i,t}^2] E[b_{j,t+p}^2 ] E [ x_{i,t}^2 x_{j,t+p}^2 ] + \sum_{i} E[ b_{i,t}^2 b_{i,t+p}^2 ] E [ x_{i,t}^2 x_{i,t+p}^2 ] + (\sigma_{e_{\pi(n+1)}}^2)^2 \\[2pt]
		& + \sum_{i \neq j} 4 E[ b_{i,t} b_{i,t+p}] E[b_{j,t} b_{j,t+p}] E [ x_{i,t} x_{i,t+p} x_{j,t} x_{j,t+p}] + \sum_{i} E[ b_{i,t}^2] E[ x_{i,t}^2] \sigma_{e_{\pi(n+1)}}^2  \\[2pt]
		&+ \sum_{i} E[ b_{i,t+p}^2] E[ x_{i,t+p}^2] \sigma_{e_{\pi(n+1)}}^2   \\[5pt]
		
		= & \sum_{i \neq j} E[ b_{i,t}^2] E[b_{j,t}^2 ] E [ x_{i,t}^2 x_{j,t+p}^2 ] + \sum_{i} \frac{1+2 \alpha_i^{2p}}{3} E[ b_{i,t}^4 ] E [ x_{i,t}^2 x_{i,t+p}^2 ]  + (\sigma_{e_{\pi(n+1)}}^2)^2 \\[2pt]
		& +  \sum_{i \neq j} 4 \alpha_i^p \alpha_j^p E[ b_{i,t}^2] E[b_{j,t}^2 ] E [ x_{i,t} x_{i,t+p} x_{j,t} x_{j,t+p}] + \sum_{i} E[ b_{i,t}^2] E[ x_{i,t}^2] \sigma_{e_{\pi(n+1)}}^2 
		+ \sum_{i} E[ b_{i,t}^2] E[ x_{i,t+p}^2] \sigma_{e_{\pi(n+1)}}^2  \\[5pt]
		
		= & \sum_{i \neq j} \frac{\sigma_i^2}{1-\alpha_i^2} \frac{\sigma_j^2}{1-\alpha_j^2} E [ x_{i,t}^2 x_{j,t+p}^2 ] + \sum_{i} \frac{1+2 \alpha_i^{2p}}{3} 3 (\frac{\sigma_i^2}{1-\alpha_i^2})^2 E [ x_{i,t}^2 x_{i,t+p}^2 ] + (\sigma_{e_{\pi(n+1)}}^2)^2 \\[2pt]
		&  +  \sum_{i \neq j} 4 \alpha_i^p \alpha_j^p \frac{\sigma_i^2}{1-\alpha_i^2} \frac{\sigma_j^2}{1-\alpha_j^2} E [ x_{i,t} x_{i,t+p} x_{j,t} x_{j,t+p}]  + \sum_{i} \frac{\sigma_i^2}{1-\alpha_i^2} E[ x_{i,t}^2] \sigma_{e_{\pi(n+1)}}^2  + \sum_{i} \frac{\sigma_i^2}{1-\alpha_i^2} E[ x_{i,t+p}^2] \sigma_{e_{\pi(n+1)}}^2 \\
		\end{array}
		\label{Stat1}
		\end{equation}
		with $1 \leq i,j \leq \pi(n)$, $E[b_{i,t}^2] = \frac{\sigma_i^2}{1-\alpha_i^2}$, and $E[b_{i,t}^4] = 3 (\frac{\sigma_i^2}{1-\alpha_i^2})^2$.
		We can see that $S(t,t+p)_{\pi(n+1)}$ is determined by corresponding parameters in the causal models of $x_{\mathbf{V}_n,t}$ and a linear combination of cross-statistics of different orders of $x_{\mathbf{V}_n,t}$.
		We denote the set of parameters by $\theta_{\pi(n+1)}$. We can find a $\theta_{\pi(n+1)}$, so that $\forall p$,  $S(t,t+p)_{\pi(n+1)}$ can be totally explained by a linear combination of cross-statistics of different orders of $x_{\mathbf{V}_n,t}$.
		
		\vspace{5cm}
		\newpage
		Next we show that for other nodes $r_{s} \in \bar{\mathbf{V}}_{n}$ which are at the $n'$th level with $n'>n+1$, $S(t,t+p)_{r_s} $ can not be totally explained by a linear combination of cross-statistics of different orders of $x_{\mathbf{V}_n,t}$.
		For $x_{\pi(n')}$, its potential causes are $x_{\mathbf{V}_n,t} \cup z$; here we use $z$ to denote the set of variables which are from the $(n+2)$th to $(n'-1)$th level. Then
		\begin{equation}
		\begin{array}{ll}
		& S(t,t+p)_{\pi(n')} \\
		= &E \big [x_{\pi(n'),t}^2 x_{\pi(n'),t+p}^2 \big] \\[5pt]
		= &E \big[ ( b^T_{\pi(n'),t} x_{\mathbf{V}_n,t} + b_{z,t}^T z_t + e_{\pi(n'),t} )^2  ( b^T_{\pi(n'),t+p} x_{\mathbf{V}_n,t+p} + b_{z,t}^T z_t + e_{\pi(n'),t+p} )^2 \big] \\[5pt]
		
		= &E[ (b^T_{\pi(n'),t} x_{\mathbf{V}_n,t})^2 (b^T_{\pi(n'),t+p} x_{\mathbf{V}_n,t+p})^2] + (\sigma_{e_{\pi(n')}}^2)^2 
		+ E[ (b^T_{\pi(n'),t} x_{\mathbf{V}_n,t})^2] \sigma_{e_{\pi(n')}}^2 \\[2pt]
		&+ E[ (b^T_{\pi(n'),t+p} x_{\mathbf{V}_n,t+p})^2] \sigma_{e_{\pi(n')}}^2 
		+ E[ (b^T_{\pi(n+1),t} x_{\mathbf{V}_n,t})^2 (b^T_{z,t+p} z_{t+p})^2 ] 
		+  E[ (b^T_{\pi(n+1),t+p} x_{\mathbf{V}_n,t+p})^2 (b^T_{z,t} z_{t})^2 ] \\[2pt]
		&  +  E[ (b^T_{z,t} z_{t})^2 ] \sigma_{e_{\pi(n')}}^2 +  E[ (b^T_{z,t+p} z_{t+p})^2 ] \sigma_{e_{\pi(n')}}^2  + E[ (b^T_{z,t} z_{t})^2 (b^T_{z,t+p} z_{t+p})^2 ] \\[2pt]
		&+ 4 E[ (b^T_{\pi(n+1),t} x_{\mathbf{V}_n,t})^2 (b^T_{\pi(n+1),t+p} x_{\mathbf{V}_n,t+p})^2 (b^T_{z,t} z_{t})^2 (b^T_{z,t+p} z_{t+p})^2 ]\\[5pt]
		
		= & \sum_{i \neq j} E[ b_{i,t}^2] E[b_{j,t}^2 ] E [ x_{i,t}^2 x_{j,t+p}^2 ] + \sum_{i} \frac{1+2 a_i^{2p}}{3} E[ b_{i,t}^4 ] E [ x_{i,t}^2 x_{i,t+p}^2 ] 
		+  \sum_{i \neq j} 4 a_i^p a_j^p E[ b_{i,t}^2] E[b_{j,t}^2 ] E [ x_{i,t} x_{i,t+p} x_{j,t} x_{j,t+p}] \\[2pt]
		&  + \sum_{i} E[ b_{i,t}^2] E[ x_{i,t}^2] \sigma_{e_{\pi(n')}}^2  + \sum_{i} E[ b_{i,t}^2] E[ x_{i,t+p}^2] \sigma_{e_{\pi(n')}}^2  + (\sigma_{e_{\pi(n')}}^2)^2 
		
		
		+ \sum_{i', j} E[ b_{i',t}^2] E[b_{j,t}^2 ] E [ x_{i',t}^2 x_{j,t+p}^2 ] \\[2pt]
		&+ \sum_{i'} \frac{1+2 a_{i'}^{2p}}{3} E[ b_{i',t}^4 ] E [ x_{i',t}^2 x_{i',t+p}^2 ]   +  \sum_{i', j} 4 a_{i'}^p a_j^p E[ b_{i',t}^2] E[b_{j,t}^2 ] E [ x_{i',t} x_{i',t+p} x_{j,t} x_{j,t+p}] \\[2pt]
		&  + \sum_{i'} E[ b_{i',t}^2] E[ x_{i',t}^2] \sigma_{e_{\pi(n')}}^2  + \sum_{i'} E[ b_{i',t}^2] E[ x_{i',t+p}^2] \sigma_{e_{\pi(n')}}^2   + \sum_{i, j'} E[ b_{i,t}^2] E[b_{j',t}^2 ] E [ x_{i,t}^2 x_{j',t+p}^2 ] \\[2pt]
		&+  \sum_{i \neq j'} 4 a_i^p a_{j'}^p E[ b_{i,t}^2] E[b_{j',t}^2 ] E [ x_{i,t} x_{i,t+p} x_{j',t} x_{j',t+p}] 
		+ \sum_{i', j'} E[ b_{i',t}^2] E[b_{j',t}^2 ] E [ x_{i',t}^2 x_{j',t+p}^2 ] \\[2pt]
		&+  \sum_{i' \neq j'} 4 a_{i'}^p a_{j'}^p E[ b_{i',t}^2] E[b_{j',t}^2 ] E [ x_{i',t} x_{i',t+p} x_{j',t} x_{j',t+p}] \\[5pt]

		= &\textcolor{black}{ \sum_{i \neq j}  \frac{\sigma_i^2}{1-\alpha_i^2}  \frac{\sigma_j^2}{1-\alpha_j^2} E [ x_{i,t}^2 x_{j,t+p}^2 ] + \sum_{i} \frac{1+2 a_i^{2p}}{3} 3 (\frac{\sigma_i^2}{1-\alpha_i^2})^2 E [ x_{i,t}^2 x_{i,t+p}^2 ] 
			+  \sum_{i \neq j} 4 a_i^p a_j^p \frac{\sigma_i^2}{1-\alpha_i^2}  \frac{\sigma_j^2}{1-\alpha_j^2} E [ x_{i,t} x_{i,t+p} x_{j,t} x_{j,t+p}] }\\[2pt]
		&  \textcolor{black}{  + \sum_{i} \frac{\sigma_i^2}{1-\alpha_i^2} E[ x_{i,t}^2] \sigma_{e_{\pi(n')}}^2  + \sum_{i} \frac{\sigma_i^2}{1-\alpha_i^2} E[ x_{i,t+p}^2] \sigma_{e_{\pi(n')}}^2  + (\sigma_{e_{\pi(n')}}^2)^2} \\[2pt]
		
		& + \sum_{i', j} \frac{\sigma_{i'}^2}{1-\alpha_{i'}^2}  \frac{\sigma_j^2}{1-\alpha_j^2} E [ x_{i',t}^2 x_{j,t+p}^2 ] + \sum_{i'} \frac{1+2 a_{i'}^{2p}}{3} 3 (\frac{\sigma_{i'}^2}{1-\alpha_{i'}^2})^2 E [ x_{i',t}^2 x_{i',t+p}^2 ]  \\[2pt]
		& +  \sum_{i', j} 4 a_{i'}^p a_j^p \frac{\sigma_{i'}^2}{1-\alpha_{i'}^2}  \frac{\sigma_j^2}{1-\alpha_j^2} E [ x_{i',t} x_{i',t+p} x_{j,t} x_{j,t+p}]   + \sum_{i'} \frac{\sigma_{i'}^2}{1-\alpha_{i'}^2} E[ x_{i',t}^2] \sigma_{e_{\pi(n')}}^2 + \sum_{i'} \frac{\sigma_{i'}^2}{1-\alpha_{i'}^2} E[ x_{i',t+p}^2] \sigma_{e_{\pi(n')}}^2  \\[2pt]
		& + \sum_{i, j'}  \frac{\sigma_i^2}{1-\alpha_i^2}  \frac{\sigma_{j'}^2}{1-\alpha_{j'}^2} E [ x_{i,t}^2 x_{j',t+p}^2 ]+  \sum_{i, j'} 4 a_i^p a_{j'}^p \frac{\sigma_i^2}{1-\alpha_i^2}  \frac{\sigma_{j'}^2}{1-\alpha_{j'}^2} E [ x_{i,t} x_{i,t+p} x_{j',t} x_{j',t+p}] \\[2pt]
		& + \sum_{i', j'} \frac{\sigma_{i'}^2}{1-\alpha_{i'}^2}  \frac{\sigma_{j'}^2}{1-\alpha_{j'}^2} E [ x_{i',t}^2 x_{j',t+p}^2 ] 
		+  \sum_{i' \neq j'} 4 a_{i'}^p a_{j'}^p \frac{\sigma_{i'}^2}{1-\alpha_{i'}^2}  \frac{\sigma_{j'}^2}{1-\alpha_{j'}^2} E [ x_{i',t} x_{i',t+p} x_{j',t} x_{j',t+p}] \\[2pt]
		\end{array}
		\label{Stat2}
		\end{equation}
		with $1 \leq i,j \leq n$, $n+1 \leq i',j' < n'$.
		Denote by  $st(x_{\mathbf{V}_n,t},\theta_{\pi(n+1)},p)$ the sum of first six terms in the last equality in (\ref{Stat2}) and $st(x_{\mathbf{V}_{n'-1,t}},\theta_{\pi(n')},p)$ the remaining parts. 
		Provided that $st(x_{\mathbf{V}_{n'-1,t}},\theta_{\pi(n')},p)$ =  $a \cdot st(x_{\mathbf{V}_n,t},\theta_{\pi(n+1)},p)$ does not hold for any $a \in \mathbf{R}$,  $S(t,t+p)_{\pi(n')}$ cannot be determined by parameters in $\theta_{\pi(n+1)}$ and statistics of $x_{\mathbf{V}_n,t}$. Thus, we can not find a set of parameters, so that $S(t,t+p)_{\pi(n')}$ can be totally explained by a linear combination of cross-statistics of different orders of $x_{\mathbf{V}_n,t}$.
		
		Thus, we can determine the process which is at the $(n+1)$th level, and its parameters are identifiable according to (\ref{DLM}) and Lemma \ref{Lemma: S1}. Therefore, we can identify the causal model up to the $(n+1)$th level.
		
		Repeating this procedure until we go through all processes, we have the identifiability of the whole causal model.

	\end{proof}

	\section{Proof of Corollary 1}
	\begin{proof}
		Since in Theorem 1, we have shown that the instantaneous causal order is identifiable, and for lagged causal relations, their causal order is fixed: from past to future, it reduces to a parameter identification problem.
		
		For variable with ordering $\pi(i)$, it has the following varying coefficients model:
		\begin{equation}
		\left \{
		\begin{array}{lll} 
		x_{\pi(i),t} = \sum \limits_{j=1}^{\pi(i-1)} b_{\pi(i)j,t} x_{j,t} + \sum  \limits_{s=1}^{s_l} \sum  \limits_{k=1}^{m} c_{\pi(i)k,t}^{(s)} x_{k,t-s} + e_{\pi(i),t},\\[2pt]
		b_{\pi(i)j,t}  = \alpha_{\pi(i)j,0} + \alpha_{\pi(i)j,1} b_{\pi(i)j,t-1} + \epsilon_{\pi(i)j,t}, \\[2pt]
		c_{\pi(i)j,t}^{(s)}  = \gamma_{\pi(i)j,0}^{(s)} + \sum\limits_{r=1} \gamma_{\pi(i)j,r}^{(s)} c_{\pi(i)j,t-r}^{(s)} + \nu_{\pi(i)j,t}^{(s)},
		\end{array}\right.
		\label{SSM2}
		\end{equation}
		where the instantaneous causal order $\pi$ is known.
		
		According to Lemma \ref{Lemma: S1}, the parameters in the varying coefficients model are identifiable. Thus, the parameters in (\ref{SSM2}) are identifiable. Combined with the result from Theorem 1 that the causal order is identifiable, therefore, the whole causal model is identifiable. 
	\end{proof}

	\section{M Step in SAEM algorithm}
	In this section, we give detailed derivations of the M step in the SAEM algorithm. We consider three scenarios separately: with the change of both causal strengths and noise variances (Section \ref{Sec: M1}), with the change of only causal strengths (Section \ref{Sec: M2}), and with both instantaneous and time-lagged causal relationships (Section \ref{Sec: M3}).
	\subsection{With the Change of Both Causal Strengths and Noise Variances} \label{Sec: M1}
	The proposed time-varying causal network is defined as:
	\begin{equation} 
	\label{SSM_supp}
	\left\{  
	\begin{array}{lll}
	X_t & = (I - B_t)^{-1} E_t, \\
	b_{ij,t} & = \alpha_{ij,0} + \sum\limits_{p=1}^{p_l} \alpha_{ij,p} b_{ij,t-p} + \epsilon_{ij,t}, \\  
	h_{i,t} & = \beta_{i,0} + \sum\limits_{q=1}^{q_l} \beta_{i q} h_{i,t-q} + \eta_{i,t},  \\
	\end{array}
	\right.
	\end{equation}
	for $t = max(p_l,q_l),\cdots,T$, where $X_t = (x_{1,t}, \cdots, x_{m,t})^{\text{T}}$, $B_t$ is an $m \times m$ causal adjacency matrix with entries $b_{ij,t}$, and $E_t = (e_{1,t}, \cdots, e_{m,t})^{\text{T}}$ with $e_{i,t} \sim \mathcal{N}(0, \sigma_{i,t}^2)$, $h_{i,t} = \log(\sigma^2_{i,t}) $, $ \epsilon_{ij,t} \sim \mathcal{N}(0,w_{ij})$, and $\eta_{i,t} \sim \mathcal{N}(0,v_i)$.
	
	The model defined in equation (\ref{SSM_supp}) can be regarded as a nonlinear state space model, with causal coefficients and the logarithm of noise variances being latent variables $Z =\big \{ \{b_{ij}\}_{ij}, \{h_i\}_i \big \}$, and model parameters $\theta = \big \{\{\alpha_{ij,0}\}_{ij}, \{\alpha_{ij,p}\}_{ij,p}, \{\beta_{i,0}\}_{i}, \{\beta_{i,q}\}_{i,q}, \{w_{ij}\}_{ij}, \{v_i\}_i\big \}$. Therefore, it can be transformed to a standard nonlinear state space model estimation. Particularly, we exploit an efficient stochastic approximation expectation maximization (SAEM) algorithm \cite{SAEM}, combining with conditional particle filters with ancestor sampling (CPF-AS) in the E step \cite{CPF, SAEM_CPF}, for model estimation.
	
	SAEM computes the E step by Monte Carlo integration and uses a stochastic approximation update of the quantity $\mathcal{Q}$:
	\begin{equation} \label{Q}
	\hat{\mathcal{Q}}_k (\theta) = (1 - \lambda_k)  \hat{\mathcal{Q}}_{k-1} (\theta) + \lambda_k  \sum_{j=1}^{M} \frac{\omega_T^{(k,j)}}{\sum_l \omega_T^{(k,l)}} \log p_{\theta} (X_{1:T}, B_{1:T}^{(k,j)}, h_{1:T}^{(k,j)}),
	\end{equation}
	with
	\begin{equation}
	\begin{array}{ll}
	& \log p_{\theta} (X_{1:T}, B_{1:T}^{(k,j)}, h_{1:T}^{(k,j)}) \\[2pt]
	= &\sum_{t=1}^{T} \log p_{\theta} (X_t | B_t^{(k,j)}, h_t^{(k,j)}) +  \sum_{t=p_l+1}^{T} \log p_{\theta} (B_t^{(k,j)} | B_{t-1}^{(k,j)}, \cdots, B_{t-p_l}^{(k,j)}) + \sum_{t=1}^{p_l} \log p_{\theta} (B_t^{(k,j)}) \\[2pt]
	& +  \sum_{t=q_l+1}^{T} \log p_{\theta} (h_t^{(k,j)} | h_{t-1}^{(k,j)},\cdots, h_{t-q_l}^{(k,j)}) + \sum_{t=1}^{q_l} \log p_{\theta} (h_t^{(k,j)}),
	\end{array}
	\end{equation}
	where $B_{t}^{(k,j)}$ is the sampled $j$th particle of $B_t$ at the $k$th iteration, and $h_{t}^{(k,j)}$ is the sampled $j$th particle of $h_t$ at the $k$th iteration.
	
	Let $\alpha_0 = \{\alpha_{ij,0}\}_{ij}$, $\alpha_p = \{\alpha_{ij,p}\}_{ij}$ for $p = 1,\cdots, p_l$, $ w = \{w_{ij}\}_{ij}$, and let $\beta_0 = \{\beta_{i,0}\}_{i}$, $\beta_q = \{\beta_{i,q}\}_{i,q}$ for $q = 1,\cdots, q_l$,  $v = \{v_i\}_i$.
	
	For presentation convenience, we reorganize the form of some parameters and latent variables. Let $\tilde{B}$ be an $m(m-1) \times 1$ vector,  which is derived by stacking each column of $B$ in sequence after removing diagonal entries.  Let $\tilde{\alpha}_p$ be an $m(m-1) \times m(m-1)$ matrix,  which is derived first by stacking each column of $\alpha$ in sequence after removing the diagonal entries and then diagonalize the vector into a matrix. The same operation is applied on $\alpha_0$ to get $\tilde{\alpha}_0$, and on $w$ to get $\tilde{w}$. $\tilde{\beta}_q$ is an $m \times m$ matrix, derived by diagonalize the vector $\beta_q$ into a matrix. The same operations are applied on $\beta_0$ to get $\tilde{\beta}_0$. The reorganized parameters are $\tilde{\theta} = \{\tilde{\alpha}_0, \{\tilde{\alpha}_p\}_{p=1}^{p_l}, \tilde{w}, \tilde{\beta}_0, \{\tilde{\beta}_q\}_{q=1}^{q_l} , v \}$.
	
	By inductive reasoning, equation $(\ref{Q})$ can be rewritten as
	\begin{equation}
	\hat{\mathcal{Q}}_k (\theta) = \sum_{i=1}^{k} \sum_{j=1}^{M} (1-\lambda_k)(1-\lambda_{k-1})\cdots (1-\lambda_{i+1}) \lambda_i \cdot \omega_T^{(i,j)}  \cdot L^{(i,j)},
	\end{equation}
	where $L^{(i,j)} = \log p_{\theta} (X_{1:T}, B_{1:T}^{(i,j)}, h_{1:T}^{(i,j)}) $. Each parameter is estimated by setting the corresponding partial derivative of the expected log-likelihood $\hat{\mathcal{Q}}_k $ to zero. 
	
	By taking the derivative of $ \hat{\mathcal{Q}}_k (\theta)$ w.r.t $\tilde{\alpha}_p$, we have
	\begin{equation}
	\begin{array}{ll}
	& \frac{\partial \hat{\mathcal{Q}}_k}{\partial \tilde{\alpha}_p} \\
	= & \frac{\partial }{\partial \tilde{\alpha}_p} \sum_{i=1}^{k} \sum_{j=1}^{M} (1-\lambda_k)(1-\lambda_{k-1})\cdots (1-\lambda_{i+1}) \lambda_i \cdot \omega_T^{(i,j)}  \cdot L^{(i,j)} \\
	= & \sum_{i=1}^{k} \sum_{j=1}^{M} (1-\lambda_k)(1-\lambda_{k-1})\cdots (1-\lambda_{i+1})(1-\lambda_i) \cdot \omega_T^{(i,j)} \cdot \frac{\partial  L^{(i,j)}}{\partial \tilde{\alpha}_p} 
	\end{array}
	\end{equation}
	with
	\begin{equation}
	\frac{\partial L^{(i,j)}}{\partial \tilde{\alpha}_p} = \sum_{t=p_l+1}^{T} \tilde{w}^{-1} (\tilde{B}_t^{(i,j)} - \tilde{\alpha}_0 - \sum_{l = 1}^{p_l}\tilde{\alpha}_l \cdot  \tilde{B}_{t-l}^{(i,j)}) \tilde{B}_{t-p}^{T (i,j)}.
	\end{equation}
	Set $\frac{\partial \hat{\mathcal{Q}}_k}{\partial \tilde{\alpha}_p}= 0$, and thus,
	\begin{equation}
	\begin{array}{ll}
	& \hat{\tilde{\alpha}}_p^{(k)} \\ 
	\! \! \! = \! \! \! & \! \! \! \! \! \! \! \! \! \bigg(  \sum_{i=1}^{k} \sum_{j=1}^{M} \sum_{t=p_l+1}^{T} (1-\lambda_k)(1-\lambda_{k-1})\cdots (1-\lambda_{i+1}) \lambda_i \cdot \omega_T^{(i,j)} \cdot (\tilde{B}_t^{(i,j)}  - \tilde{\alpha}_0 - \sum_{l \neq p}\tilde{\alpha}_l \cdot  \tilde{B}_{t-l}^{(i,j)} )\tilde{B}_{t-p}^{T (i,j)}   \bigg) \\
	& \cdot  \bigg(  \sum_{i=1}^{k} \sum_{j=1}^{M} \sum_{t=p_l+1}^{T} (1-\lambda_k)(1-\lambda_{k-1})\cdots (1-\lambda_{i+1}) \lambda_i \cdot \omega_T^{(i,j)} \cdot  \tilde{B}_{t-p}^{(i,j)} \tilde{B}_{t-p}^{T (i,j)}   \bigg)^{-1}\\
	= &\underbrace{ \bigg((1-\lambda_k) \hat{\tilde{\alpha}}_{p_a}^{(k-1)} + \lambda_k  \sum_{j=1}^{M} \sum_{t=p_l+1}^{T}  \omega_T^{(k,j)} \cdot (\tilde{B}_t^{(k,j)}  - \tilde{\alpha}_0 - \sum_{l \neq p}\tilde{\alpha}_l \cdot  \tilde{B}_{t-l}^{(k,j)} )\tilde{B}_{t-p}^{T (k,j)}     \bigg)}_{\hat{\tilde{\alpha}}_{p_a}^{(k)}}\\
	&  \cdot \underbrace{ \bigg((1-\lambda_k) \hat{\tilde{\alpha}}_{p_b}^{(k-1)} + \lambda_k  \sum_{j=1}^{M} \sum_{t=p_l+1}^{T}  \omega_T^{(k,j)} \cdot \tilde{B}_{t-p}^{(k,j)} \tilde{B}_{t-p}^{T (k,j)}     \bigg)^{-1} }_{\hat{\tilde{\alpha}}_{p_b}^{(k)}}.
	\end{array}
	\end{equation}

	Set $\frac{\partial \hat{\mathcal{Q}}_k}{\partial \tilde{\alpha}_0} = 0$, and thus 
	\begin{equation}
	\begin{array}{ll}
	& \hat{\tilde{\alpha}}_0^{(k)}\\
	= & \bigg(\sum_{i=1}^{k} \sum_{j=1}^{M} (T-p_l) \cdot (1-\lambda_k)(1-\lambda_{k-1})\cdots (1-\lambda_{i+1}) \lambda_i \cdot \omega_T^{(i,j)} \bigg)^{-1} \\ 
	& \cdot \bigg( \sum_{i=1}^{k} \sum_{j=1}^{M} \sum_{t=p_l+1}^{T} (1-\lambda_k)(1-\lambda_{k-1})\cdots (1-\lambda_{i+1}) \lambda_i \cdot \omega_T^{(i,j)} \cdot (B_t^{(i,j)} - \tilde{\alpha}_0 -  \sum_{l=1}^{p_l} \tilde{\alpha}_l \cdot B_{t-l}^{(i,j)})   \bigg) \\
	= & \underbrace{ \bigg((1-\lambda_k) \hat{\alpha}_{0_a}^{(k-1)} + \lambda_k  \sum_{j=1}^{M} (T-p_l) \cdot  \omega_T^{(k,j)}     \bigg)^{-1}}_{\hat{\alpha}_{0_a}^{(k)}}\\
	& \cdot \underbrace{ \bigg((1-\lambda_k) \hat{\alpha}_{0_b}^{(k-1)} + \lambda_k  \sum_{j=1}^{M} \sum_{t=p_l+1}^{T}  \omega_T^{(k,j)} \cdot (B_t^{(k,j)} - \tilde{\alpha}_0 -  \sum_{l=1}^{q_l} \tilde{\alpha}_l \cdot B_{t-l}^{(k,j)})     \bigg) }_{\hat{\alpha}_{0_b}^{(k)}}.
	\end{array}
	\end{equation}

	Take the derivative of $ \hat{\mathcal{Q}}_k (\theta)$ w.r.t $\tilde{w}$:               
	\begin{equation}
	\frac{\partial \hat{\mathcal{Q}}_k}{\partial \tilde{w} } 
	= \sum_{i=1}^{k} \sum_{j=1}^{M} (1-\lambda_k)(1-\lambda_{k-1})\cdots (1-\lambda_{i+1}) \lambda_i \cdot \omega_T^{(i,j)} \cdot \frac{\partial L^{(i,j)}}{\partial \tilde{w} } ,
	\end{equation}
	where
	\begin{equation}
	\begin{array}{ll}
	& \frac{\partial L^{(i,j)}}{\partial \tilde{w} } \\
	= & \sum_{t=p_l+1}^{T} \big[  -\frac{1}{2} \tilde{w}^{-T} + \frac{1}{2}\tilde{w}^{-T} (\tilde{B}_t^{(i,j)} - \tilde{\alpha}_0 -  \sum_{l = 1}^{p_l}\tilde{\alpha}_l \cdot  \tilde{B}_{t-l}^{(i,j)})  (\tilde{B}_t^{(i,j)} - \tilde{\alpha}_0 - \sum_{l = 1}^{p_l}\tilde{\alpha}_l \cdot  \tilde{B}_{t-l}^{(i,j)})^T \tilde{w}^{-T} \big].
	\end{array}
	\end{equation}
	Set $\frac{\partial \hat{\mathcal{Q}}_k}{\partial \tilde{w} } = 0$, and thus
	\begin{equation}
	\begin{array}{ll}
	& \hat{\tilde{w}}^{(k)}\\
	= &\bigg(\sum_{i=1}^{k} \sum_{j=1}^{M} (T-p) \cdot (1-\lambda_k)(1-\lambda_{k-1})\cdots (1-\lambda_{i+1}) \lambda_i \cdot \omega_T^{(i,j)} \bigg)^{-1} \\ 
	& \cdot \bigg( \sum_{i=1}^{k} \sum_{j=1}^{M} \sum_{t=p+1}^{T} (1-\lambda_k)(1-\lambda_{k-1})\cdots (1-\lambda_{i+1}) \lambda_i \cdot \omega_T^{(i,j)} \\
	& \quad \cdot (\tilde{B}_t^{(i,j)} - \tilde{\alpha}_0 - \sum_{l = 1}^{p_l}\tilde{\alpha}_l \cdot  \tilde{B}_{t-l}^{(i,j)})  (\tilde{B}_t^{(i,j)} - \tilde{\alpha}_0 - \sum_{l = 1}^{p_l}\tilde{\alpha}_l \cdot  \tilde{B}_{t-l}^{(i,j)})^T  \bigg) \\
	= &\underbrace{ \bigg((1-\lambda_k) \hat{\tilde{w}}_a^{(k-1)} + \lambda_k  \sum_{j=1}^{M} (T-p) \cdot  \omega_T^{(k,j)}     \bigg)^{-1}}_{\hat{\tilde{w}}_a^{(k)}}\\
	&  \cdot \underbrace{ \bigg((1-\lambda_k) \hat{\tilde{w}}_b^{(k-1)} + \lambda_k  \sum_{j=1}^{M} \sum_{t=p_l+1}^{T}  \omega_T^{(k,j)} \cdot (\tilde{B}_t^{(k,j)} - \tilde{\alpha}_0 - \sum_{l = 1}^{p_l}\tilde{\alpha}_l \cdot  \tilde{B}_{t-l}^{(k,j)})  (\tilde{B}_t^{(k,j)} - \tilde{\alpha}_0 -  \sum_{l = 1}^{p_l}\tilde{\alpha}_l \cdot  \tilde{B}_{t-l}^{(k,j)})^T     \bigg) }_{\hat{\tilde{w}}_b^{(k)}}.
	\end{array}
	\end{equation}

	Take the derivative of $ \hat{\mathcal{Q}}_k (\theta)$ w.r.t $\beta_0$ and set
	$\frac{\partial \hat{\mathcal{Q}}_k}{\partial \beta_0} = 0$, and thus 
	\begin{equation}
	\begin{array}{ll}
	& \hat{\beta}_0^{(k)}\\
	= & \bigg(\sum_{i=1}^{k} \sum_{j=1}^{M} (T-q) \cdot (1-\lambda_k)(1-\lambda_{k-1})\cdots (1-\lambda_{i+1}) \lambda_i \cdot \omega_T^{(i,j)} \bigg)^{-1} \\ 
	& \cdot \bigg( \sum_{i=1}^{k} \sum_{j=1}^{M} \sum_{t=q+1}^{T} (1-\lambda_k)(1-\lambda_{k-1})\cdots (1-\lambda_{i+1}) \lambda_i \cdot \omega_T^{(i,j)} \cdot (h_t^{(i,j)} - \tilde{\beta}_0 -  \sum_{l=1}^{q_l} \tilde{\beta}_l \cdot h_{t-l}^{(i,j)})   \bigg) \\
	= & \underbrace{ \bigg((1-\lambda_k) \hat{\beta}_{0_a}^{(k-1)} + \lambda_k  \sum_{j=1}^{M} (T-q) \cdot  \omega_T^{(k,j)}     \bigg)^{-1}}_{\hat{\beta}_{0_a}^{(k)}}\\
	& \cdot \underbrace{ \bigg((1-\lambda_k) \hat{\beta}_{0_b}^{(k-1)} + \lambda_k  \sum_{j=1}^{M} \sum_{t=q_l+1}^{T}  \omega_T^{(k,j)} \cdot (h_t^{(k,j)} - \tilde{\beta}_0 -  \sum_{l=1}^{q_l} \tilde{\beta}_l \cdot h_{t-l}^{(k,j)})     \bigg) }_{\hat{\beta}_{0_b}^{(k)}}.
	\end{array}
	\end{equation}
	
	Take the derivative of $ \hat{\mathcal{Q}}_k (\theta)$ w.r.t $\tilde{\beta}_q$ and 
	set $\frac{\partial \hat{\mathcal{Q}}_k}{\partial \tilde{\beta}_q}= 0$, and thus
	\begin{equation}
	\begin{array}{ll}
	& \hat{\tilde{\beta}}_q^{(k)} \\ 
	=  & \bigg(  \sum_{i=1}^{k} \sum_{j=1}^{M} \sum_{t=q_l}^{T} (1-\lambda_k)(1-\lambda_{k-1})\cdots (1-\lambda_{i+1}) \lambda_i \cdot \omega_T^{(i,j)} \cdot (h_t^{(i,j)} - \beta_0 - \sum_{l \neq q} \beta_l h_{t-l}^{(i,j)}) h_{t-q}^{T (i,j)}   \bigg) \\
	& \quad \cdot  \bigg(  \sum_{i=1}^{k} \sum_{j=1}^{M} \sum_{t=q_l}^{T} (1-\lambda_k)(1-\lambda_{k-1})\cdots (1-\lambda_{i+1}) \lambda_i \cdot \omega_T^{(i,j)} \cdot h_{t-q}^{(i,j)}  h_{t-q}^{T (i,j)} \bigg)^{-1}\\
	= & \underbrace{ \bigg((1-\lambda_k) \hat{\tilde{\beta}}_{q_a}^{(k-1)} + \lambda_k  \sum_{j=1}^{M} \sum_{t=q_l+1}^{T}  \omega_T^{(k,j)} \cdot       (h_t^{(k,j)} - \beta_0 - \sum_{l \neq q} \beta_l h_{t-l}^{(k,j)}) h_{t-q}^{T (k,j)}    \bigg)}_{\hat{\tilde{\beta}}_{q_a}^{(k)}}\\
	&  \cdot \underbrace{ \bigg((1-\lambda_k) \hat{\tilde{\beta}}_{q_b}^{(k-1)} + \lambda_k  \sum_{j=1}^{M} \sum_{t=q_l+1}^{T}  \omega_T^{(k,j)} \cdot h_{t-q}^{(k,j)}  h_{t-q}^{T (k,j)}   \bigg)^{-1} }_{\hat{\tilde{\beta}}_{q_b}^{(k)}}.
	\end{array}
	\end{equation}

	Take the derivative of $ \hat{\mathcal{Q}}_k (\theta)$ w.r.t $v$ and 
	set $\frac{\partial \hat{\mathcal{Q}}_k}{\partial v } = 0$, and thus
	\begin{equation}
	\begin{array}{ll}
	& \hat{v}^{(k)}\\
	= &\bigg(\sum_{i=1}^{k} \sum_{j=1}^{M} (T-q_l) \cdot (1-\lambda_k)(1-\lambda_{k-1})\cdots (1-\lambda_{i+1}) \lambda_i \cdot \omega_T^{(i,j)} \bigg)^{-1} \\ 
	& \cdot \bigg( \sum_{i=1}^{k} \sum_{j=1}^{M} \sum_{t=q_l+1}^{T} (1-\lambda_k)(1-\lambda_{k-1})\cdots (1-\lambda_{i+1}) \lambda_i \cdot \omega_T^{(i,j)} \\
	& \qquad \cdot (h_t^{(i,j)} - \beta_0 -  \sum_{l=1}^{q_l} \tilde{\beta}_l h_{t-l}^{(i,j)}) (h_t^{(i,j)} - \beta_0 -  \sum_{l=1}^{q_l} \tilde{\beta}_l h_{t-l}^{(i,j)}) ^T  \bigg) \\
	= & \underbrace{ \bigg((1-\lambda_k) \hat{v}_a^{(k-1)} + \lambda_k  \sum_{j=1}^{M} (T-q_l) \cdot  \omega_T^{(k,j)}     \bigg)^{-1}}_{\hat{v}_a^{(k)}}\\
	&  \cdot \underbrace{ \bigg((1-\lambda_k) \hat{v}_b^{(k-1)} + \lambda_k  \sum_{j=1}^{M} \sum_{t=q_l+1}^{T}  \omega_T^{(k,j)} \cdot    (h_t^{(k,j)} - \beta_0 -  \sum_{l=1}^{q_l} \tilde{\beta}_l h_{t-l}^{(k,j)}) (h_t^{(k,j)} - \beta_0 -  \sum_{l=1}^{q_l} \tilde{\beta}_l h_{t-l}^{(k,j)}) ^T     \bigg) }_{\hat{v}_b^{(k)}}.
	\end{array}
	\end{equation}
	

	\subsection{With the Change of Only Causal Strengths} \label{Sec: M2}
	If we assume that the variance of $E_t$ does not change across time, then we have the following time-varying causal network:
	\begin{equation} 
	\left\{  
	\begin{array}{lllll}
	X_t  & =  & (I - B_t)^{-1} E_t, \\
	b_{ij,t} & = & \alpha_{ij,0} + \sum\limits_{p=1}^{p_l} \alpha_{ij,p} b_{ij,t-p} + \epsilon_{ij,t} , \\  
	\end{array}
	\right.
	\end{equation}
	for $t = p_l,\cdots,T$, where $X_t = (x_{1,t}, \cdots, x_{m,t})^{\text{T}}$, $B_t$ is an $m \times m$ causal adjacency matrix with entries $b_{ij,t}$, $E_t = (e_{1,t}, \cdots, e_{m,t})^{\text{T}}$ with $e_{i,t} \sim \mathcal{N}(0, \sigma_{i}^2)$, and $ \epsilon_{ij,t} \sim \mathcal{N}(0,w_{ij})$. Let $R$ be an $m \times m$ diagonal matrix with diagonal entries $\sigma_i^2$. The latent variables are $Z =\big \{ \{b_{ij}\}_{ij}\big \}$, and model parameters are $\theta = \big \{\{\alpha_{ij,0}\}_{ij}, \{\alpha_{ij,p}\}_{ij,p}, \{w_{ij}\}_{ij},  \{\sigma_i\}_i\big \}$. 
	
	We update $R$ by setting $\frac{\partial \hat{\mathcal{Q}}_k}{\partial \tilde{R} } = 0$, and thus
	\begin{equation}
	\begin{array}{ll}
	& \hat{R}^{(k)}\\
	= & \bigg(\sum_{i=1}^{k} \sum_{j=1}^{M} T \cdot (1-\lambda_k)(1-\lambda_{k-1})\cdots (1-\lambda_{i+1}) \lambda_i \cdot \omega_T^{(i,j)} \bigg)^{-1} \\ 
	& \cdot \bigg( \sum_{i=1}^{k} \sum_{j=1}^{M} \sum_{t=1}^{T} (1-\lambda_k)(1-\lambda_{k-1})\cdots (1-\lambda_{i+1}) \lambda_i \cdot \omega_T^{(i,j)} \cdot (I -  \tilde{B}_{t}^{(i,j)}) X_t  X_t^T (I - \tilde{B}_{t}^{(i,j)})^T  \bigg) \\
	= &\underbrace{ \bigg((1-\lambda_k) \hat{\tilde{R}}_a^{(k-1)} + \lambda_k  \sum_{j=1}^{M} T \cdot  \omega_T^{(k,j)}     \bigg)^{-1}}_{\hat{\tilde{R}}_a^{(k)}}\\
	&  \cdot \underbrace{ \bigg((1-\lambda_k) \hat{\tilde{R}}_b^{(k-1)} + \lambda_k  \sum_{j=1}^{M} \sum_{t=1}^{T}  \omega_T^{(k,j)} \cdot (I -  \tilde{B}_{t}^{(k,j)}) X_t  X_t^T (I - \tilde{B}_{t}^{(k,j)})^T     \bigg) }_{\hat{\tilde{R}}_b^{(k)}}.
	\end{array}
	\end{equation}
	
	The other parameters are updated in the same way as those is Section \ref{Sec: M1}.
	
	\subsection{With Both Contemporaneous and Time-Lagged Causal Relationships} \label{Sec: M3}
	It is easy to extend to the case with both contemporaneous and time-lagged causal relationships:
	\begin{equation}
	\left \{
	\begin{array}{lllll} 
	X_t &= & (I-B_t)^{-1} (\sum_{s}^{s_l} C_t^{(s)} X_{t-s} + E_t), \\
	b_{ij,t}  &=& \alpha_{ij,0} + \sum_{p=1}^{p_l} \alpha_{ij,p} b_{ij,t-p} + \epsilon_{ij,t},  \\
	h_{i,t} & = & \beta_{i,0} + \sum\limits_{q=1}^{q_l} \beta_{i q} h_{i,t-q} + \eta_{i,t}, \\ 
	c_{ij,t}^{(s)}  &=& \gamma_{ij,0}^{(s)} + \sum\limits_{r=1}^{r_l} \gamma_{ij,r}^{(s)} c_{ij,t-r}^{(s)} + \nu_{ij,t}^{(s)}, \\
	\end{array}\right.
	\end{equation}
	for $t = max(p_l,q_l),\cdots,T$, where $X_t = (x_{1,t}, \cdots, x_{m,t})^{\text{T}}$, $B_t$ is an $m \times m$ causal adjacency matrix with entries $b_{ij,t}$, $C_t^{(s)}$ is an $m \times m$ matrix with entries $c_{ij,t}^{(s)}$ for $s = 1, \cdots,s_l$, and $E_t = (e_{1,t}, \cdots, e_{m,t})^{\text{T}}$ with $e_{i,t} \sim \mathcal{N}(0, \sigma_{i,t}^2)$, $h_{i,t} = \log(\sigma^2_{i,t}) $, $ \epsilon_{ij,t} \sim \mathcal{N}(0,w_{ij})$, $\eta_{i,t} \sim \mathcal{N}(0,v_i)$, and $\nu_{ij,t}^{(s)} \sim \mathcal{N}(0,u_{ij}^{(s)})$. 
	The latent variables are $Z =\big \{ \{b_{ij}\}_{ij}, \{c_{ij}^{(s)}\}_{ij,s}, \{h_i\}_i \big \}$, and model parameters are $\theta = \big \{\{\alpha_{ij,0}\}_{ij}, \{\alpha_{ij,p}\}_{ij,p}, \{\beta_{i,0}\}_{i}, \{\beta_{i,q}\}_{i,q}, \{w_{ij}\}_{ij}, \{v_i\}_i ,   \{\gamma_{ij,0}^{(s)}\}_{ij,s}, \{\gamma_{ij,r}^{(s)}\}_{ij,r,s}, \{u_{ij}\}_{ij}^{(s)} \big \}.$

	Set $\frac{\partial \hat{\mathcal{Q}}_k}{\partial \tilde{\gamma}_r^{(s)}} = 0$, and thus
	\begin{equation}
	\begin{array}{ll}
	& \hat{\tilde{\gamma}}_r^{(s,k)} \\ 
	& =  \bigg(  \sum_{i=1}^{k} \sum_{j=1}^{M} \sum_{t=r_l+1}^{T} (1-\lambda_k)(1-\lambda_{k-1})\cdots (1-\lambda_{i+1}) \lambda_i \cdot \omega_T^{(i,j)} \cdot (\tilde{C}_t^{(s,i,j)} - \gamma_0^{(s)} - \sum_{l \neq r} \gamma_{l}^{(s)} \tilde{C}_{t-l}^{(s,i,j)}) \tilde{C}_{t-r}^{T (s,i,j)}    \bigg) \\
	& \quad \cdot  \bigg(  \sum_{i=1}^{k} \sum_{j=1}^{M} \sum_{t=r_l+1}^{T} (1-\lambda_k)(1-\lambda_{k-1})\cdots (1-\lambda_{i+1}) \lambda_i \cdot \omega_T^{(i,j)} \cdot \tilde{C}_{t-r}^{(s,i,j)} \tilde{C}_{t-r}^{T (s,i,j)}   \bigg)^{-1}.\\
	& =\underbrace{ \bigg((1-\lambda_k) \hat{\tilde{\gamma}}_{r_a}^{(s,k-1)} + \lambda_k  \sum_{j=1}^{M} \sum_{t=r_l+1}^{T}  \omega_T^{(k,j)} \cdot (\tilde{C}_t^{(s,k,j)} - \gamma_0^{(s)} - \sum_{l \neq r} \gamma_{l}^{(s)} \tilde{C}_{t-l}^{(s,k,j)}) \tilde{C}_{t-r}^{T (s,k,j)}     \bigg)}_{\hat{\tilde{\gamma}}_{r_a}^{(s,k)}}\\
	& \quad \cdot \underbrace{ \bigg((1-\lambda_k) \hat{\tilde{\gamma}}_{r_b}^{(s,k-1)} + \lambda_k  \sum_{j=1}^{M} \sum_{t=r_l+1}^{T}  \omega_T^{(k,j)} \cdot \tilde{C}_{t-r}^{(s,k,j)} \tilde{C}_{t-r}^{T (s,k,j)}     \bigg)^{-1} }_{\hat{\tilde{\gamma}}_{r_b}^{(s,k)}}
	\end{array}
	\end{equation}

	Set $\frac{\partial \hat{\mathcal{Q}}_k}{\partial \tilde{u}^{(s)} } = 0$, and thus
	\begin{equation}
	\begin{array}{ll}
	& \hat{\tilde{u}}^{(s,k)}\\
	= &\bigg(\sum_{i=1}^{k} \sum_{j=1}^{M} (T-r_l) \cdot (1-\lambda_k)(1-\lambda_{k-1})\cdots (1-\lambda_{i+1}) \lambda_i \cdot \omega_T^{(i,j)} \bigg)^{-1} \\ 
	& \cdot \bigg( \sum_{i=1}^{k} \sum_{j=1}^{M} \sum_{t=r_l+1}^{T} (1-\lambda_k)(1-\lambda_{k-1})\cdots (1-\lambda_{i+1}) \lambda_i \cdot \omega_T^{(i,j)} \\
	& \quad \cdot (\tilde{C}_t^{(s,i,j)} - \tilde{\gamma}_0^{(s)} - \sum_{l = 1}^{r_l}\tilde{\gamma}_l \cdot  \tilde{C}_{t-l}^{(s,i,j)})  (\tilde{C}_t^{(s,i,j)} - \tilde{\gamma}_0^{(s)} - \sum_{l = 1}^{p_l}\tilde{\gamma}_l \cdot  \tilde{C}_{t-l}^{(s,i,j)})^T  \bigg) \\
	= &\underbrace{ \bigg((1-\lambda_k) \hat{\tilde{u}}_a^{(s,k-1)} + \lambda_k  \sum_{j=1}^{M} (T-r_l) \cdot  \omega_T^{(k,j)}     \bigg)^{-1}}_{\hat{\tilde{u}}_a^{(s,k)}}\\
	&  \cdot \underbrace{ \bigg((1-\lambda_k) \hat{\tilde{u}}_b^{(s,k-1)} + \lambda_k  \sum_{j=1}^{M} \sum_{t=r_l+1}^{T}  \omega_T^{(k,j)} \cdot (\tilde{C}_t^{(s,k,j)} - \tilde{\gamma}_0^{(s)} - \sum_{l = 1}^{r_l}\tilde{\gamma}_l^{(s)} \cdot  \tilde{C}_{t-l}^{(s,k,j)})  (\tilde{C}_t^{(s,k,j)} - \tilde{\gamma}_0^{(s)} -  \sum_{l = 1}^{p_l}\tilde{\gamma}_l^{(s)} \cdot  \tilde{C}_{t-l}^{(s,k,j)})^T     \bigg) }_{\hat{\tilde{u}}_b^{(s,k)}}.
	\end{array}
	\end{equation}
	
	Set $\frac{\partial \hat{\mathcal{Q}}_k}{\partial \tilde{u}_c } = 0$, and thus
	\begin{equation}
	\begin{array}{ll}
	& \hat{\tilde{u}}_c^{(k)}\\
	= & \bigg(\sum_{i=1}^{k} \sum_{j=1}^{M} (1-\lambda_k)(1-\lambda_{k-1})\cdots (1-\lambda_{i+1}) \lambda_i \cdot \omega_T^{(i,j)} \bigg)^{-1} \\ 
	& \cdot \bigg( \sum_{i=1}^{k} \sum_{j=1}^{M} (1-\lambda_k)(1-\lambda_{k-1})\cdots (1-\lambda_{i+1}) \lambda_i \cdot \omega_T^{(i,j)} \cdot \frac{1}{r_l} \sum_{t=1}^{r_l}(\tilde{C}_t^{(i,j)} - \mu_c)  (\tilde{C}_t^{(i,j)} - \mu_c)^T  \bigg)\\
	= &\underbrace{ \bigg((1-\lambda_k) \hat{\tilde{u}}_{c_a}^{(k-1)} + \lambda_k  \sum_{j=1}^{M}  \omega_T^{(k,j)}     \bigg)^{-1}}_{\hat{\tilde{u}}_{c_a}^{(k)}}\\
	& \cdot \underbrace{ \bigg((1-\lambda_k) \hat{\tilde{u}}_{c_b}^{(k-1)} + \lambda_k  \sum_{j=1}^{M}  \omega_T^{(k,j)} \cdot \frac{1}{r_l} \sum_{t=1}^{r_l} (\tilde{C}_t^{(k,j)} - \mu_c)  (\tilde{C}_t^{(k,j)} - \mu_c)^T     \bigg) }_{\hat{\tilde{u}}_{c_b}^{(k)}}.
	\end{array}
	\end{equation}

	Set $\frac{\partial \hat{\mathcal{Q}}_k}{\partial \tilde{\gamma}_0^{(s)}} = 0$, and thus 
	\begin{equation}
	\begin{array}{ll}
	& \hat{\tilde{\gamma}}_0^{(s,k)}\\
	= & \bigg(\sum_{i=1}^{k} \sum_{j=1}^{M} (T-r_l) \cdot (1-\lambda_k)(1-\lambda_{k-1})\cdots (1-\lambda_{i+1}) \lambda_i \cdot \omega_T^{(i,j)} \bigg)^{-1} \\ 
	& \cdot \bigg( \sum_{i=1}^{k} \sum_{j=1}^{M} \sum_{t=r_l+1}^{T} (1-\lambda_k)(1-\lambda_{k-1})\cdots (1-\lambda_{i+1}) \lambda_i \cdot \omega_T^{(i,j)} \cdot (C_t^{(s,i,j)} - \tilde{\gamma}_0^{(s)} -  \sum_{l=1}^{r_l} \tilde{\gamma}_l^{(s)} \cdot C_{t-l}^{(s,i,j)})   \bigg) \\
	= & \underbrace{ \bigg((1-\lambda_k) \hat{\tilde{\gamma}}_{0_a}^{(s,k-1)} + \lambda_k  \sum_{j=1}^{M} (T-r_l) \cdot  \omega_T^{(k,j)}     \bigg)^{-1}}_{\hat{\tilde{\lambda}}_{0_a}^{(s,k)}}\\
	& \cdot \underbrace{ \bigg((1-\lambda_k) \hat{\tilde{\gamma}}_{0_b}^{(s,k-1)} + \lambda_k  \sum_{j=1}^{M} \sum_{t=q_l+1}^{T}  \omega_T^{(k,j)} \cdot (C_t^{(s,k,j)} - \tilde{\gamma}_0^{(s)} -  \sum_{l=1}^{q_l} \tilde{\gamma}_l^{(s)} \cdot C_{t-l}^{(s,k,j)})     \bigg) }_{\hat{\tilde{\gamma}}_{0_b}^{(s,k)}}.
	\end{array}
	\end{equation}
	
	The other parameters are updated with the same way as those in Section \ref{Sec: M1}.

	\section{Conditional Particle Filter with Ancestor Sampling}

	The detailed procedure of conditional particle filter with ancestor sampling is summarized in Algorithm \ref{CPF-AS}.
	\begin{algorithm}
		\caption{CPF with Ancestor Sampling}
		\begin{algorithmic}[1]
			\STATE Let prespecified particles be $\mathring{Z}'_{1:T} = \{\mathring{Z}'_1, \cdots, \mathring{Z}'_T\}$.
			\STATE Draw $\mathring{Z}_1^{(j)}$ from $\mathring{Z}_1^{(j)} \sim f_{\theta}(\mathring{Z}_1), \text{ } j = 1,\cdots,M-1$.
			\STATE Set $\mathring{Z}_1^{(M)} = \mathring{Z}'_1$.
			\STATE Set $\omega_1^{(j)} = W_{\theta,1}(\mathring{Z}_1^{(j)})$ for $j= 1,\cdots,M$.
			\FOR {$t = 2 \text{ to } T$} 
			\STATE  Draw $s_t^j$ with $P(s_t^j = i) \propto \omega_{t-1}^{(i)}$ for $j = 1,\cdots,M-1$.
			\STATE Draw $s_t^M$ with $P(s_t^M = j) \propto \omega_{t-1}^{(j)} f_{\theta}(\mathring{Z}'_t | \mathring{Z}_{t-1}^j)$.
			\STATE Draw $\mathring{Z}_t^{(j)} \sim f_{\theta}(\mathring{Z}_t | \mathring{Z}_{t-1}^{s_t^j})$ for $\text{ } j = 1,\cdots,M-1$.
			
			\STATE Set $\mathring{Z}_t^{(M)} = \mathring{Z}'_t$.
			\STATE  Set $\mathring{Z}_{1:t}^{(j)} = \{\mathring{Z}_{1:t-1}^{s_t^j}, \mathring{Z}_t^{(j)}\}$ for $j = 1,\cdots,M$.
			\STATE Set $\omega_t^{(j)} = W_{\theta,t}(\mathring{Z}_t^{(j)}, \mathring{Z}_{t-1}^{s_t^j})$ for $j = 1,\cdots,M$.
			\ENDFOR
		\end{algorithmic}	
		\label{CPF-AS}	   	
	\end{algorithm}

	\section{Sparsity Constraints}
	In practical problems, the causal connections may be sparse. In this section, we consider the sparsity constraints on causal adjacency matrix $B_t$ and that on $b_{ij,t} - b_{ij,t-1}, \forall i, j$, which ensures smooth changes of $b_{ij,t}$ across time.
	
	It is well known that lasso regularization for sparsity is biased. Thus, we utilize the smoothly clipped absolute deviation (SCAD) penalty, which has shown to be unbiased for the resulting estimator of significant parameters \cite{SCAD}. The SCAD penalty is given by
	\begin{flalign*} 
	p_{\lambda}^{\text{SCAD}}(b_{ij,t}) = 
	\left\{  
	\begin{array}{llll}  
	\lambda |b_{ij,t}| & \text{ if } |b_{ij,t}| \leq \lambda\\  
	-\frac{|b_{ij,t}|^2-2a \lambda |b_{ij,t}| + \lambda^2}{2(a-1)} & \text{ if } \lambda < |b_{ij,t}| \leq a \lambda \\
	\frac{(a+1) \lambda^2}{2} & \text{ if } |b_{ij,t}| > a \lambda 
	\end{array}  
	\right. 
	\end{flalign*} 
	where $a$ and $\lambda$ are hyperparameters.
	The penalized log-likelihood for equation is modified as 
	\begin{equation}
	\begin{array}{ll}
	& 	p_{\lambda}(X| B) \\        [2pt]
	= & \sum_{t=1}^{T} \log p(X_t| B_t, \sigma^2_t)  - \sum_{t=1}^{T}  \sum_{i,j} p_{\lambda}^{\text{SCAD}}(b_{ij,t})   \\      [2pt]
	& -  \sum_{t=2}^{T} \sum_{i,j}p_{\lambda}^{\text{SCAD}}(b_{ij,t}-b_{ij,t-1}).\\ [2pt]
	\end{array}
	\end{equation}
	
	Since adding sparsity on $B_t$ and $b_{ij,t} - b_{ij,t-1}$ does not affect the derivatives of the parameters in the M step, we only need to modify the likelihood used in the E step, by replacing $p(X| B)$ with $p_{\lambda}(X| B)$.

\bibliography{term_fin1}
\bibliographystyle{icml2019}

\end{document}